\documentclass[11pt]{article}

\usepackage{acl}
\usepackage{latexsym}
\usepackage{microtype}
\usepackage{xurl}  

\usepackage{graphicx}
\usepackage{amsmath}
\usepackage{amssymb}
\usepackage{booktabs}
\usepackage{float}
\usepackage[section]{placeins}  
\usepackage{longtable}
\usepackage{array}
\usepackage{fontspec}
\usepackage{multirow}

\newfontfamily\kannadafont{NotoSansKannada.ttf}

\title{Cross Lingual Transfer in Tulu Legal Comprehension: Script-Dependent Improvement and RAG-Induced Knowledge Conflict}

\author{Sindhu Shetty, Spurthi Setty, Natan Vidra}

\begin{document}
\maketitle

\begin{abstract}

Low-resource languages without an adequate training corpus often use a related, higher-resource language as a scaffold for comprehension. Still, there is a need to develop rigorous evaluation methods to identify when models fail in cross lingual low-resource environments. Using the legal domain as a backdrop, three models (Llama3, Hex-1, Sarvam) were tested on the ability to classify legal complaints written in a low resource Dravidian language (Tulu). Transliterating queries across Dravidian scripts allowed models to gain a preliminary understanding of speakers' complaints without the use of wide scale training, though the level of comprehension was heavily script dependent (with Kannada - another relatively low-resource language - producing the strongest positive trend). Retrieving from a corpus of Kannada legal papers across a RAG framework caused mixed results. Some models had a weak positive trend in comprehension under certain conditions, but when models failed, it was often across two axes: fact substitution (fixating on specific passage excerpts that skewed reasoning) and confabulation (hallucination that had no basis in either query or corpus). Within low resource domains, results identify the model's parsing of information and subsequent reasoning as the source of reasoning failure, rather than corpus contents. Script-dependent comprehension and RAG robustness also seem to travel together. This is further supported by the reasoning-trace analysis and a statistical-honesty framework deployed - techniques that are more broadly applicable to low-resource multilingual RAG evaluation.

\end{abstract}

\section{Introduction}
Despite the various capabilities of large language models, one of the most pressing pitfalls is how models assess information in low-resource domains. In many regions of the world, the common language spoken (often a highly low-resource language) differs from the administrative language. To exacerbate the issue, the administrative language is often low-resource itself, requiring  novel techniques for models to bridge this cross-lingual low-resource domain. A relevant application of this dilemma is the legal field, where members of such communities often need to represent themselves in a non-native language to access resources.

This study uses the Dravidian language Tulu, a small language with a few million speakers off the coast of Karnataka. The official language of Karnataka is Kannada, meaning all legal proceedings take place with Kannada or English mediums. Kannada is itself classified a low resource language, as despite having an extensive literary history, training LLM on it has only picked up recently. Tulu is extremely low resource; it does not contain the extensive literature used to traditionally train models.

With a low resource language, Tulu, how would it be possible to file speakers' complaints into a system that does not administratively use their language (and instead uses Kannada), without a translator? In this study, we explore whether minimal priming and legal papers written in Kannada are enough for a model to accurately interpret queries written in a language they do not have previous exposure to. 

While high-resource languages are often used as a starting point when working with low-resource languages, the intuition points in the other direction. When Tulu is written, it has traditionally used Kannada abugidas rather than the Latin alphabet. Tulu speakers are also more likely to borrow vocabulary and structure sentences similarly to the more common language around them - Kannada. Thus, we propose that use of the Kannada (and other Dravidian) scripts would trigger stronger comprehension in models. Furthermore, legal papers written in Kannada would provide a stronger foothold for an LLM to understand Tulu complaints spoken in an informal register due to the aforementioned relation between the two languages. 

The purpose of this paper is not to create a full, working LLM for low resource languages, but to evaluate model comprehension in a cross-lingual low-resource domain, using the legal field as an appropriate, grounded testing ground. To do so, we ask three LLM models to categorize legal complaints written in Tulu using the Legal Issues Taxonomy (LIST) from the Stanford Legal Design Lab \citep{stanfordlegaldesignlab_list}, without translation or heavy training. We explore whether minimal prompting and RAG from a corpus in a higher resource, related language, can serve as a scaffold for models to pinpoint the gaps in their understanding of the complaint. 

Most Tulu speakers are multilingual and can communicate in Kannada, English, Hindi, etc, though doing so can still result in the loss of nuance that occurs when speaking in a second language. 

\section{Related Work}
\textbf{Knowledge Conflict} - When training large language models, they often run into knowledge conflict, as shown by \citep{longpre2021entity}. When models were forced to decide between a newly given passage or their own memory, they often fall back to their own understanding, even when prompted otherwise. However, later findings diverge from this understanding. \citep{xu2024knowledge} establishes the "context-memory conflict", where models struggle to properly weigh retrieved information that conflicts with their own internal understanding, and can inconsistently defer to context that is coherent and confidently stated, regardless of accuracy. This is further explored in \citep{wu2024clasheval}, which found that models consistently abandon their correct prior knowledge in favor of retrieved context - including Llama3 - one of the models used in this study. \citep{xie2024adaptive} lends credence to the claim that models are receptive to confidently presented external evidence, even if a knowledge conflict is present, though notes another separate confirmation bias - models tend to stick with their own prior understanding when the evidence set only partially agrees with it. Given that model comprehension in a low resource, legal context is still not widely understood, our unique angle isolates how these mechanisms can be further utilized by comparing multiple models' innate ability, retrieval analysis, and discernment between the two in a multilingual setting. 

\textbf{Effect of RAG Inclusion} - RAG's inclusion can be detrimental to knowledge synthesis. \citep{hsia2024ragged} explains how various model/retriever combinations can affect downstream performance, sometimes performing worse compared to no retrieval at all. Witnessed as well in \citep{zeng2025ragguard}, where models performed worse when presented with misleading retrieved evidence compared to a zero shot baseline and could not recognize when to discard incorrect information. Our work furthers this discussion, displaying how models are able to use minimal priming to make their predictions more accurate, but the inclusion of RAG prevents most models from effectively discerning when to disregard irrelevant passages. 

\textbf{Hallucination in Legal NLP} - It's especially vital to understand this issue in legal work, where small knowledge conflicts can snowball into major hallucinations. \citep{magesh2025hallucination} establishes how commercial legal AI built around RAG still meaningfully hallucinates, with literature like \citep{reuter2025towards} working towards mitigation. Our findings add to established literature, demonstrating that even relevant legal papers can cause hallucination within a model, which may be due to cross-lingual transfer or uncertainty in a low resource environment.

\textbf{Retrieval and Low Resource Languages} - This problem is magnified among low resource languages: \citep{sun2026exposing} illustrate how low-resource language pairs rely more heavily on retrieved context and degrade more severely when that context is noisy compared to high-resource pairs. \citep{hu2025crosslingual} extends this, demonstrating that while knowledge-free reasoning easily transfers across language despite the secondary resource effect, knowledge retrieval does not. This may explain why RAG frequently failed when the query was not written in a high resource language, which did not allow the model to discriminate between irrelevant passages and its own reasoning. Our methods also draw from \citep{qi2025consistency}, which separates context utilization from retrieval quality, though ours differ by maintaining consistent retrieval over queries written in multiple languages. Our findings are also supported by \citep{mohan2026retrieval}, which explains how a failure occurs not from retrieved content itself, but how the model uses it - rationale for why introducing retrieved passages often failed, as the issue may lie with the model's ingestion of the passages as well as confusion with multilingual transfer across a low resource domain. This intersection of low resource multilingual transfer across the legal domain is relatively unexplored. 

\textbf{Tulu-Kannada Relation} - Within the scope of the Tulu-Kannada bridge, \citep{madasamy2024tulu} establishes that because of the extensive contact between the two languages, using Kannada as a scaffold for Tulu seems like a viable option. This is supported by \citep{devadiga2026making}, which discovers how Tulu is "nearly indistinguishable" from Kannada to Kannada trained LLM. Our results align, given the drop in comprehension when Tulu is written in other Dravidian scripts compared to Kannada.

\section{Methodology}

\subsection{Task Formulation}
To discover to what extent a model could understand queries in Tulu, we authored 60 sentences describing common legal situations, with each sentence ascribed a primary category, inspired by the Legal Issues Taxonomy (LIST) from the Stanford Legal Design Lab \citep{stanfordlegaldesignlab_list} (where some categories were replaced), as well as any relevant secondary categories. The model had to use the sentence given (as well as any additional context) to predict what category the Tulu complaint fell under, with performance evaluated across the 60 sentences. The hope was to build a pipeline that could classify complaints in a low resource language with minimal prompting.

\subsection{Data and Models}
\label{subsec:data-models}
The 60 Tulu sentences were written in an informal, common register describing a situation requiring legal analysis. The sentences were written in Latin script as well as three Indic scripts, primarily Kannada, but also Malayalam and Tamil, and translated into English as a control.

The legal corpus consisted of 18 legal documents from Karnataka courts (taken from ecourtsindia.com), written in Kannada. Documents were orders of various sizes, ranging from a paragraph and a table describing common proceedings to 25-page hearings detailing witness testimonies. Many corpus documents consisted of images or used legacy non-Unicode fonts that produced mojibake when extracted (garbled Latin text rather than standard Kannada), requiring automated text extraction. To clean up extraction based errors, AI assisted correction was utilized, an established technique \citep{kanerva2025ocr}. Using Gemini, original documents were given alongside the scraped text for the model to identify and correct text based errors, ensuring the model had a reference to source material rather than open-ended inference. 

The cleaned legal papers were each ascribed to at least one major LIST category, although secondary categories could be ascribed as well. The documents exist solely for the model to gain an understanding of similar problems to the query, to aid diagnosis.

Three models were used to understand cross-lingual legal transfer. Llama3 \citep{meta2024llama3} had no official support for Kannada or Tulu, though a small portion (over 5\%) of its training data covered more than 30 additional languages not officially documented. Sarvam \citep{sarvamai_models} and Hex-1 \citep{budecosystem2025hex1} were both trained with a focus on Indic languages including Kannada, but not Tulu, testing whether a Kannada basis helps comprehension. All three models had temperature 0.1 for reproducibility; Llama3 and Hex-1 had a repeat penalty of 1.1. Hex-1's context window (num\_ctx) was raised to 8192 to prevent truncation. Enabling thinking mode on Hex-1 caused inner reasoning with no response, while Sarvam's reasoning consumed all tokens in its "starter tier" (4000 tokens), also yielding no output, so thinking/reasoning was disabled for both. 

For embedding the legal texts, we originally used nomic-embed-text, but switched to multilingual-e5-large-instruct, as the latter produced embeddings spanning a wider range rather than deferring to a default median.

\subsection{Experimental Conditions}
\label{subsec:experimental-conditions}
Each condition was composed of four parts: (1) the script and language the query was written in, (2) whether the model received linguistic priming, (3) whether the model received RAG passages from the Kannada legal corpus, and (4) if primed, whether the priming was sparse or extensive.

Priming had two variants. \textbf{Sparse} priming contained a short paragraph of common Tulu vocabulary (most not present in the test sentences) and basic grammatical structure. \textbf{Extensive} priming was a continuation of sparse priming that additionally included words related to describing legal situations - e.g. family, land, permission - some of which were used directly in the 60 test sentences.

Tulu words in the priming block matched the query's script (e.g. a Romanized-Tulu query received Romanized-Tulu priming vocabulary), while instructions remained in English throughout. The English control never received priming.

Retrieved passages shown to the model included the source case name, retrieval similarity score, and the passage's own LIST category label. Baselines included English-gloss queries with RAG (Condition A), to test how RAG affected comprehension with a high-resource query language, and Romanized/Kannada-script queries without priming or RAG (Condition G and Condition H, respectively), to test latent understanding without modification.

\begin{table}[H]
\centering
\footnotesize
\setlength{\tabcolsep}{4pt}
\begin{tabular}{@{}lccc@{}}
\toprule
\textbf{Condition} & \textbf{Script} & \textbf{Priming} & \textbf{RAG} \\
\midrule
A & English gloss     & --         & \checkmark \\
B & Romanized Tulu    & --         & \checkmark \\
C & Romanized Tulu    & \checkmark & \checkmark \\
D & Kannada script    & \checkmark & \checkmark \\
E & Kannada script    & \checkmark & --         \\
F & Kannada script    & --         & \checkmark \\
G & Romanized Tulu    & --         & --         \\
H & Kannada script    & --         & --         \\
I & Tamil script      & --         & --         \\
J & Malayalam script  & --         & --         \\
K & Tamil script      & \checkmark & --         \\
L & Malayalam script  & \checkmark & --         \\
M & Romanized Tulu    & \checkmark & --         \\

\bottomrule
\end{tabular}
\caption{Experimental conditions, defined by query script, presence of linguistic priming, and use of retrieval-augmented generation (RAG). Conditions with priming (C, D, E, K, L, M) were each run at both sparse and extensive tiers.}
\label{tab:conditions}
\end{table}

\subsubsection{Prompting}
Each model was first given the query, after which retrieved RAG passages were provided if applicable. If RAG was used, the model was instructed not to classify the passages, but to use them to understand the situation described in the query. The model was again given the query, then presented a third time with explicit instructions not to classify retrieved passages. This repeated instruction was necessary, as otherwise the model often resolved to classify its interpretation of retrieved passages. 

\subsection{Retrieval Pipeline}
All legal passages were first broken into 800 character chunks, which were embedded using nomic-embed-text, then switched to multilingual-e5-large-instruct (the query was embedded similarly). The embedded query was compared against chunks and the top 3 chunks were retrieved using cosine similarity. Mean centering was briefly deployed to test whether it increased the number of relevant passages retrieved, but provided no recognizable optimization. 

\subsection{Diagnostic Methodology}
To assess whether the model was able to comprehend the query, its response was analyzed across four separate axes. 

\textbf{Tertiary Classification} - Since queries could often relate to multiple categories, each query was given two other related LIST labels, alongside its primary LIST classification. The model, after being provided the query and possible subsequent material, had to provide a primary, secondary, and tertiary LIST category, and could not repeat categories. If the primary category provided by the model matched the assigned primary classification, it received a point to its Hits@1 score. If the model's primary classification did not match, but the subsequent guesses corresponded with any of the three categories assigned to the query, it received a point to the Hits@3 score. 

A high Hits@1 and Hits@3 score indicated the model was able to successfully assign the proper LIST category, as well as any tangentially related categories. A high Hits@3 but low Hits@1 score was seen as the model partially understanding the query, but not enough to capture the nuance described. A low Hits@1 and Hits@3 score meant the model did not have strong comprehension of the queries. 

\textbf{Reasoning Trace Verification} - Models were instructed to provide reasoning for their primary guess, to ensure whether the model had actually understood the situation, or produced an unrelated reason that coincidentally aligned with the query. A model with a high number of such situations was seen as partially parsing the query, but then hallucinating the gaps in its comprehension.

\textbf{Statistical Honesty} - To ensure that discrepancies between conditions were significant and not due to random error, Fisher's Exact Test was deployed (e.g. comparing Condition E and Condition D for a given model). Binomial tests were also used, comparing a model's D-condition performance against random guessing (12.5\% accuracy) and majority class guessing (25\% accuracy, assigning the most common category across queries). Multiple runs for a specific model/condition pair were pooled for stronger statistical analysis. The p value must be less than the standard threshold of 0.05 to be statistically significant. Findings that qualified as significant were described with confident language (e.g. "demonstrates", "shows"), while those that did not were viewed as directional or indicative of a trend, even if raw percentage gaps seemed large.

\subsection{Preliminary Mitigation}
After the discovery of RAG hurting a model's discernment, a tentative prototype was tested. 

\textbf{CRAG-Style Relevance Verification } - A secondary judge layer was added before the prompt, in which the model discerned whether each retrieved passage was relevant with a binary judgment (for 20 complaints only). Each call had two forms: one required the model to use queries written in Kannada script, the other used the English (Condition A) queries. Each passage was reviewed via independent calls. Non-relevant passages were excluded, and if all passages were deemed irrelevant, the prompt fell back to the No-RAG format. This was tested on the Llama3 D conditions (RAG inclusive \& priming).

\section{Findings}
\label{sec:findings}

\begin{figure}[h]
\centering
\includegraphics[width=0.9\linewidth]{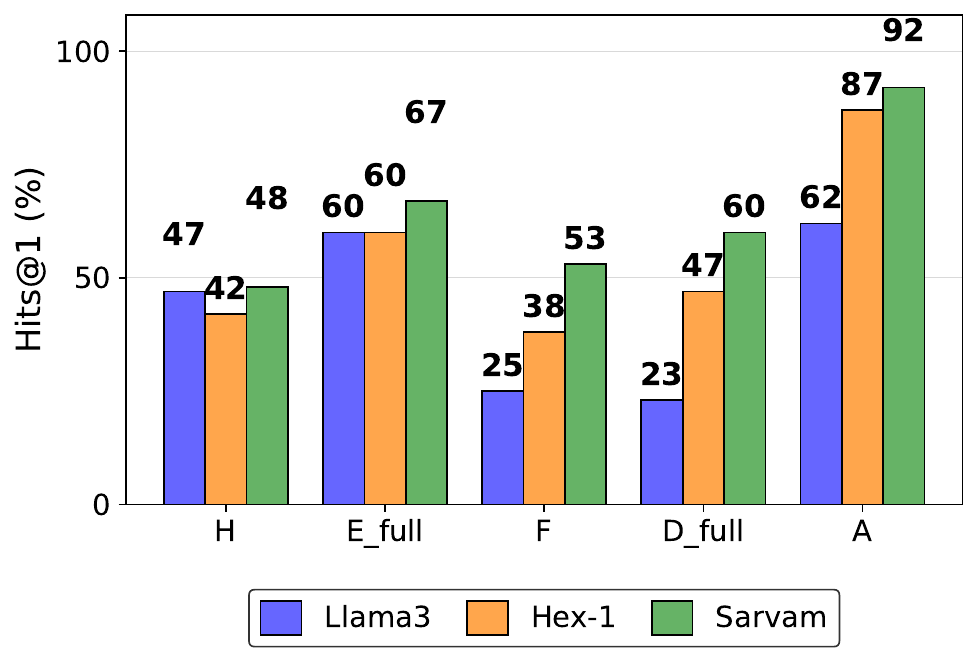}
\caption{Hits@1 across five conditions per model ($n=60$ each): bare comprehension (H), no-RAG full-priming ceiling (E\_full), RAG without priming (F), RAG with full priming (D\_full), English-gloss query with RAG (A). Only Llama3 drops below its own no-RAG ceiling once RAG is added. Full 19-condition breakdown: Appendix~\ref{app:full-findings}.}
\label{fig:findings-condensed}
\end{figure}

\begin{figure}[h]
\centering
\includegraphics[width=0.9\linewidth]{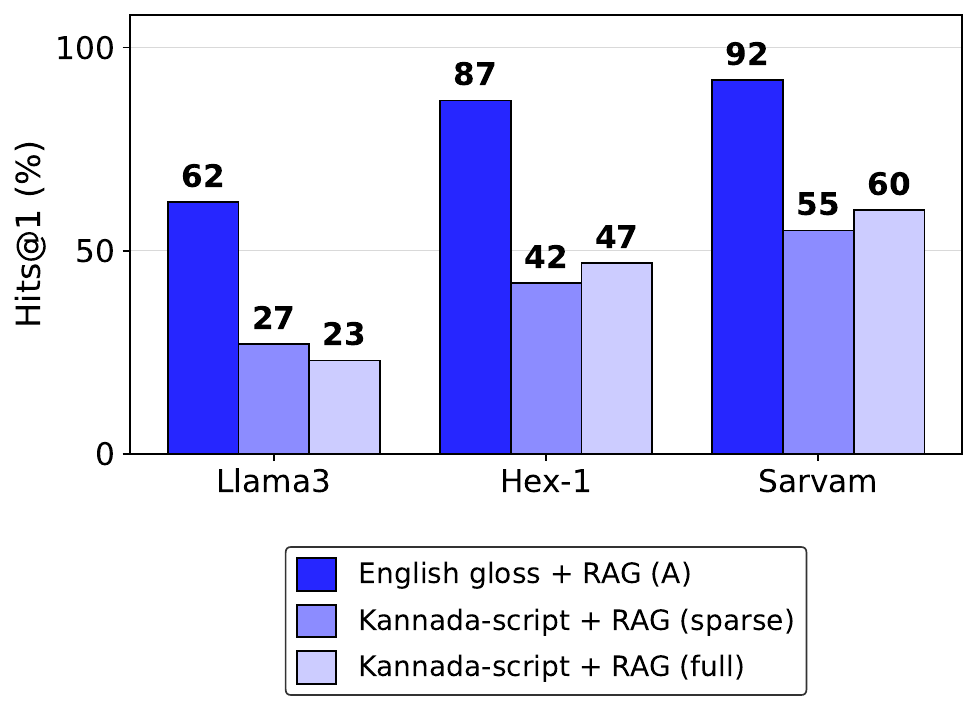}
\caption{Query language under RAG: English-gloss queries (Condition\ A) substantially outperform Kannada-script Tulu queries under identical retrieval (Condition\ D), for all three models (Fisher's exact, $p<.001$ in every comparison). This is the one RAG-related effect that generalizes across models.}
\label{fig:query-lang-rag}
\end{figure}

Across all three models, Kannada-script bare comprehension (Condition\ H) is significantly above both chance and majority-class baselines (Llama3 46.7\%, Hex-1 41.7\%, Sarvam 48.3\%; $p<.001$ in every case), and full-tier priming reveals a trend with comprehension being increased further for every model, when using Kannada (Figure~\ref{fig:findings-condensed}). The picture changes sharply once RAG is introduced. For Llama3, RAG collapses performance well below its own no-RAG ceiling (D\_full 23.3\% vs.\ E\_full 60.0\%, $p<.001$); Hex-1 and Sarvam show no significant degradation on the same comparison ($p=.20$ and $p=.57$, respectively). The full 19-condition, four-script, three-model breakdown and complete significance tests are reported in Appendix~\ref{app:full-findings}.

The one RAG-related effect that generalizes across all three models is query language: English-gloss queries under RAG (Condition A) significantly outperform Kannada-script Tulu queries under otherwise identical retrieval (Condition D/Full), for every model (p < .001 in all three comparisons; Figure \ref{fig:query-lang-rag}). A full comparison of Kannada and Romanized script performance across all priming and RAG conditions is reported in Table ~\ref{tab:kannada-romanized}. This suggests that whatever degrades performance under RAG interacts more consistently with query language than with model identity.

\section{Discussion}

\subsection{Comprehension Transfers Without RAG}
\label{subsec:comprehension_transfer}
One pressing finding was the repeatable success of Condition\ E. This trend emerges with Condition\ H, where models were able to classify given sentences without any scaffolding (at 46.7\%, 41.7\%, and 48.3\% for Llama3, Hex-1, and Sarvam respectively). In other cases, the model was able to partially understand the sentence, but not enough to accurately diagnose it. To ensure this was not random chance, we also checked the reasoning given.




\begin{table}[H]
\centering
\small
\begin{tabular}{@{}p{0.44\linewidth}p{0.44\linewidth}@{}}
\toprule
\textbf{Query (gloss)} & \textbf{Reasoning trace} \\
\midrule
``I don't know which court I should go to'' & ``The situation explicitly mentions `court', indicating a problem related to the legal system.'' \\
\addlinespace
``Someone is living on my land without asking and won't leave'' & ``...classic housing/trespass issue.'' \\
\bottomrule
\end{tabular}
\caption{Sarvam, Condition H, ids 14 and 17. Row 1: human chosen and predicted category both Courts/Legal-System, reasoning names the specific recognized term rather than a generic paraphrase. Row 2: human chosen - Land/Property, predicted - Housing. The model parses the core action but assigns an adjacent category, indicating partial rather than absent comprehension.}
\label{tab:h-examples}
\end{table}

Still, this was a preliminary indication that models were somehow able to occasionally parse queries correctly, despite no priming or previous training with the language.

Once the model received priming (Condition\ E), Llama3's and Hex-1's understanding and subsequent diagnosis improved slightly (46.7\%$\to$51.7\% and 41.7\%$\to$45.0\%, respectively, though neither reaches significance: $p=.715$ and $p=.854$), and this is not reflected with Sarvam (48.3\%$\to$46.7\%, $p=1$).

We had two theories as to how the model understood Tulu with brief exposure to the language's structure and no previous interactions with complete sentences in the target language.

The first proposed that the model was parsing Tulu as a form of Kannada, since the shared history between the two languages facilitated heavy word borrowing.\footnote{See \cite{devadiga2026making}, who find Kannada-script Tulu ``nearly indistinguishable from Kannada'' to models trained predominantly on the latter.} The second theory asserts that the model is reading Tulu as a unique Dravidian language, using features shared across the Dravidian family to make sense of the language. Given that Tulu exists in the same language family as Tamil and Malayalam (as well as Kannada), this also seemed viable.

Given the results of Condition\ I-L, script matters, though not substantially at the sparse-priming tier, and this holds across all three models. Sparse structural priming, grammar and word order only with no category-revealing vocabulary, produced only modest, non-significant changes in Kannada accuracy for every model (Llama3 46.7\%$\to$51.7\%, $p=.715$; Hex-1 41.7\%$\to$45.0\%, $p=.854$; Sarvam 48.3\%$\to$46.7\%, $p=1$), and left Tamil (Llama3 16.7\%$\to$15.0\%; Hex-1 28.3\%$\to$33.3\%; Sarvam 28.3\%$\to$33.3\%; all $p>.26$) and Malayalam (Llama3 21.7\%$\to$20.0\%; Hex-1 25.0\%$\to$28.3\%; Sarvam 21.7\%$\to$38.3\%; all $p>.07$) similarly flat or only weakly positive. If priming were teaching models portable Tulu grammar, this improvement should not be script-dependent. Since the difference does not reach significance for any script or model, this is only suggestive of script-specific assimilation over broad cross-Dravidian transfer. This effect was comparable with Romanized Tulu (Llama3 16.7\%$\to$13.3\%, $p=1$; Hex-1 16.7\%$\to$18.3\%, $p=1$; Sarvam 33.3\%$\to$30.0\%, $p=.845$).

It is undeniable that models do have latent capability for understanding queries in this low-resource language, although the workings are still unclear. Parsing through a related, higher-resource language that the target language is lexically (rather than grammatically) similar to, and specifically through matching script, seems likely, although the relatively strong comprehension among other Dravidian scripts cannot be disputed.

\subsection{RAG Causes a Large, Reproducible Collapse or No Significant Improvement}
\label{subsec:rag-collapse}
Llama3's comprehension dropped when presented with RAG, illustrated by comparing Condition\ D to F (F 25.0\% vs.\ D\_sparse 26.7\%, $p=1$; vs.\ D\_full 23.3\%, $p=1$), where priming did not help under RAG as it did in Condition\ E (H 46.7\%$\to$E\_full 60.0\%), though this did not reach significance. The issue also arises in the control (Condition\ A): even when the query is written in a high-resource language, a comparatively lower-resource retrieval corpus still degraded the model's own understanding of the query: Llama3 reaches only 61.7\% in this condition, well short of Hex-1's 86.7\% and Sarvam's 91.7\% under the identical retrieval setup. Meanwhile, both Sarvam and Hex-1 were able to maintain similar scores with the inclusion of RAG compared to just priming (Hex-1: F 38.3\% vs.\ E\_sparse 45.0\%, $p=.579$), and priming seemed to hurt Sarvam (Conditions\ E\_sparse vs.\ F: 46.7\% vs.\ 53.3\%, $p=.584$), though not significantly. This may be due to Sarvam and Hex-1 being trained on Kannada, allowing stronger discernment between passages and their own reasoning, which may also explain Llama3's heavier drop in Condition\ A.

A disagreement-level diagnostic further confirms this, which determined the ratio of Condition E (non-RAG/priming) vs Condition D (RAG/priming) producing the right answer when disagreement occurred. D failed more often E on almost all comparisons when disagreement occurred  - as high as 6.5:1 with Llama3 at full priming - except Sarvam and sparse priming in which this trend reverses (0.4:1), with the full Table~\ref{tab:disagreement-ratio} reflecting model robustness under RAG. 

When failure due to RAG did occur, the idea that the cause was knowledge conflict, not retrieval noise, is supported across several dimensions.

A model's garbled post-RAG responses fell into two categories: fact substitution and confabulation. With \textbf{fact substitution}, the model takes a specific but wrong detail present in the retrieved passage and incorporates it directly into its reasoning, treating the snippet as part of the query rather than supporting evidence. With \textbf{confabulation}, the model hallucinates information present in neither the query nor the retrieved passage, using it as a basis for reasoning regardless. This occurs even when a genuinely relevant passage is retrieved: for two debt-related queries, the retrieved case (\textit{Sridhar P.\ v.\ Ramesh}, see Appendix~\ref{app:cases}) was a real, on-topic cheque-bounce dispute, yet the model still produced unrelated categories for both.

Methodological note: retrieved passages were shown to the model together with their own source category label. A follow-up ablation stripping this label from the retrieved context produced no consistent change in accuracy (deltas within $\pm$5pp across all five RAG conditions, no consistent direction), indicating label-copying was not the primary driver of the collapse; the failure is substantially content-level, not a result of label exposure.

\subsection{Relation of RAG Inclusion and Script Across Model Performance}
While the independent effects of RAG inclusion and script have been stated, the intersection between the two highlights a possible relation - primarily evident between Kannada script and Latin script queries. With non-RAG conditions, Kannada script significantly outperformed Latin script across majority of model and condition combinations Table~\ref{tab:kannada-romanized}, lending more credence to the discussion present in \S\ref{subsec:comprehension_transfer}. Models can and do exhibit comprehension of low resource languages that they had previously never encountered, but the inclusion of Latin script adds a complicating factor - that models struggle to parse queries in a low-resource language if the script the language is written in is far removed from related languages in the model's training base (though comprehension can still exist). 

However, this difference in comprehension between script is not universally present with RAG inclusion. Sarvam and Hex-1 - both models that upheld performance with RAG inclusion (with comparison between D and E conditions)- were also models that could maintain the Kannada script advantage with RAG conditions (across Conditions C and D in Table~\ref{tab:kannada-romanized}). Llama3 - the model in which accuracy decreased (Condition D vs E), was also the only model that could not significantly beat the Latin query performance present in Condition C (though the raw numbers of Condition D were still greater). Whether a model is able to successfully overcome knowledge conflict and utilize the script advantage seem to be correlated. Furthermore, whether a model is able to uphold performance with RAG inclusion is also an indicator of whether script advantage will remain under the same RAG conditions. 

It should be noted that while Llama3's performance dropped - resulting in insignificance - Sarvam's performance was remarkably high in Latin script queries (once again causing insignificance). So while an insignificant increase in three conditions exists with Sarvam (Condition E/(no RAG \& sparse priming), Condition H (no RAG \& no priming), and Condition D/(RAG \& full priming), the underlying cause differed from Llama3. It's possible that Sarvam's explicit training in Kannada allowed it to understand some of the Latin script queries. 

Why and when a model is able to understand a script with non-RAG conditions but not under RAG conditions is still unclear. Both Sarvam and Hex-1 are trained in Kannada, which may allow them to maintain script advantage even alongside complex legal passages. Since parsing Tulu queries as Kannada produces more accurate results, the ability to uphold the script advantage could also cause the model to overcome the knowledge conflict present in RAG conditions, hence why neither model was significantly affected by the inclusion of RAG. Llama3 - the one model not explicitly trained in Kannada - may not have a strong enough basis to recognize the query language and corpus language to be similar even with the same script, hence the drop in accuracy. However, this is entirely speculative, and more work must be done to understand this phenomenon.

\subsection{Mitigation of RAG Failure}
\label{sec:mitigation}

Using Corrective Retrieval-Augmented Generation (CRAG), models tested with Kannada-script queries performed similarly to the unmitigated RAG condition (25.0\% Hits@1, vs. 23.3\% for D/full). There was an improvement (to 60.0\%) when the judge operated on English-translated complaints instead of Kannada-script queries, matching Llama3's own no-RAG ceiling (E/full, 60.0\%), even though RAG was still used in 17 of 20 cases; this difference does not clear significance at the standard threshold (p = .054, Fisher's exact), though it sits at the boundary. The English-gloss judge also discerned more passages to be irrelevant on average (1.65 of 3 retained, vs. 2.85 of 3 for the Kannada judge with full results in Table~\ref{tab:crag}), suggesting the judge layer's ability to discriminate relevant from irrelevant retrieved content depends on the language it operates in, not solely on the underlying passage relevance.

\section{Conclusion}
In communities where the spoken language differs from the administrative language, training legal models on the latter does not address the issue of communities needing to use these tools in a cross lingual manner in courts, testimonies, etc. Common languages spoken in the affected area (Tulu, in this case) may not even have a literary history, making large scale training of models difficult. With the language pair of Kannada (administrative) and Tulu (spoken), the two share a long history of word borrowing and a similar syntactical structure. Considering this basis, is it possible to bridge the gap between two low resource languages, and alleviate the difficulty of switching between vernacular and formal legal speak with RAG and minimal priming alone? The strategy seems viable, although it heavily depends on the model used. 

Given the task of classifying Tulu queries written in an informal register into specific LIST categories, models could be given two forms of priming, RAG retrieved passages, both priming and the inclusion of RAG, or neither technique at all. Surprisingly, all three models (Llama-3, Hex-1, and Sarvam) displayed relatively high levels of comprehension of the queries even with no priming or RAG at all, and both extensive and sparse priming produced an improvement over bare comprehension within conditions utilizing Kannada script for most conditions and models, though this effect did not reach significance for any model in Kannada script. However, RAG had the opposite effect on Llama3. Despite preliminary evidence that Tulu is parsed as Kannada by LLMs, the inclusion of actual Kannada legal papers only served to confuse the model on two fronts: fact substitution and confabulation. This may be because the model, treating Tulu as a form of Kannada, overly trusted retrieved passages over its own reasoning. Meanwhile, Sarvam and Hex-1 were not negatively or positively affected by RAG inclusion, though the same two mechanisms still appeared. Why Sarvam and Hex-1 did not produce the same RAG induced collapse as Llama3 is still unclear, though comparison of performance between Romanized and Kannada script queries signals that latent comprehension between scripts and the effect of RAG inclusion are interrelated rather than independent phenomena. While Llama-3 did exhibit comprehension with Kannada script, this effect did not significantly manifest once RAG was included.  Hex-1 was  able to maintain the advantage of Kannada script under both RAG and non-RAG conditions as well as Sarvam at most tiers, beating performance under Romanized queries. It is possible that due to training on Kannada, both Sarvam and Hex-1 were able to use internal knowledge and treat external sources as confirmation rather than a definitive answer. Further work must be done to understand this trend.

Llama3 (the only model hallucination reduction attempts were studied on) was somewhat responsive to the CRAG-style mitigation approach. CRAG Relevance Verification indicated the model could more easily discern irrelevant passages when the query was written in English, though this effect did not transfer to Kannada script. 

Research in multi-lingual legal transfer that does not require heavy training is necessary, as affected communities often do not have the means to do so. In future work, we hope to continue bridging this gap, exploring cross-Dravidian transfer, larger datasets, more extensive RAG hallucination mitigation methods, and different language pairs to isolate the mechanisms and make such techniques generalizable to the greatest number of communities affected by this issue. 

\section*{Limitations}

Due to the time intensive need to carefully review queries to ensure native-like production, accurate grammar, and correct transliteration, it was considered unrealistic to generate an extensive number of queries for a pilot study. This study originally employed only 20 queries to test comprehension, but this number was deemed to be too small to be able to confidently assess results.  Therefore, the final number of 60 queries was used to draw preliminary conclusions, although many of the findings were indicative of trends rather than statistically significant behavior, and would require a number of queries that enters the hundreds or thousands range to be considered significant. However, for large effects, such as the drop in Llama3 comprehension in condition D, these reach statistical significance even with a small dataset. Even so, if this study was to be expanded, having a larger number of queries would be a top priority. 

The hallucination findings of this study do raise the question of whether similar risk was present during corpus creation, as final documents were scanned but not thoroughly reviewed by a native speaker.  Kanerva et al. \citep{kanerva2025ocr} similarly found LLM-based OCR correction to be highly effective for a well-supported language (English) while not reaching such levels in a lower-resource language (Finnish). It should be noted that Gemini officially supports Kannada \citep{google2024geminiindia}. Additionally, the task required character verification rather than semantic deduction, distinguishing this corpus-creation context from that failure case. We still note this does not fully eliminate the residual risk discussed above and maintaining accuracy in future studies requiring a similar corpus should be a priority. 

As for the legal documents used themselves, their public availability raises some concerns. Specifically, there is no way to verify if the models used in this study had been previously trained on said documents. Performance under RAG-inclusive conditions could be confounded if models were previously exposed to documents present in the corpus, and the non-RAG/RAG divide discussed in this paper should be revisited if this is found to be the case. 

With regards to models, Sarvam and Hex-1 had their reasoning/thinking modes disabled, as compared to Llama3 which was allowed unconstrained free-form generation. Therefore, Sarvam's and Hex-1's responses should be analyzed knowing that they were not produced under equivalent conditions. 

\section*{Ethics Statement}

While all 60 queries are authored to represent realistic hypothetical scenarios, they are not real individuals' complaints and do not draw from real case intake data. We believe this is the best way to maintain accurate analysis without the need for private individuals' legal situations. Meanwhile, corpus documents were directly retrieved from ecourtsindia.com, where court orders are publicly accessible. Retrieved passages were not anonymized in this study. Court orders were only used to extract document content and LIST category relevant information, and personal information was never used as a basis for classification. We acknowledge this as a limitation and recommend that named-entity redaction be utilized before using retrieved passages in further studies. 

It should be heavily noted that this work is not deployment ready, given our findings of model hallucination present throughout experimentation. Deploying such pipelines has high chances of directing clients to LIST categories irrelevant to their query. Furthermore, due to the identified phenomena of fact substitution and confabulation, models can confidently provide incorrect reasoning, which can further mislead clients who do not have extensive experience in the legal field.   

While prompt construction and priming methodology are detailed in Appendix~\ref{app:glossary} and Appendix~\ref{app:prompt}, all scripts used during experimentation will be accessible in their entirety upon publication, as well as the 60 authored sentences. As for the legal corpus, the redistribution rights of the papers remain unclear, although case identifiers for examples discussed in Appendix~\ref{app:cases} are available upon request. 

\textbf{}

\bibliography{custom}

\begin{thebibliography}{20}
\providecommand{\natexlab}[1]{#1}

\bibitem[{{AI@Meta}(2024)}]{meta2024llama3}
{AI@Meta}. 2024.
\newblock \href {https://arxiv.org/abs/2407.21783} {The llama 3 herd of
  models}.
\newblock \emph{Preprint}, arXiv:2407.21783.

\bibitem[{{Bud Ecosystem}(2025)}]{budecosystem2025hex1}
{Bud Ecosystem}. 2025.
\newblock Introducing hex-1: A fully open-source {LLM} for indic languages.
\newblock
  \url{https://blog.budecosystem.com/hex1-open-source-llm-for-indic-languages/}.

\bibitem[{{Deccan Herald}(2024)}]{google2024geminiindia}
{Deccan Herald}. 2024.
\newblock Google's gemini {AI} app now available in india, supports 9
  languages.
\newblock
  \url{https://www.deccanherald.com/technology/googles-gemini-ai-app-now-available-in-india-supports-9-languages-3070891}.
\newblock Confirms official Kannada support; cited as a news source, not a
  Google primary document.

\bibitem[{Devadiga and Chopra(2026)}]{devadiga2026making}
Prathamesh Devadiga and Paras Chopra. 2026.
\newblock \href {https://doi.org/10.18653/v1/2026.loreslm-1.5} {Making large
  language models speak tulu: Structured prompting for an extremely
  low-resource language}.
\newblock In \emph{Proceedings of the Second Workshop on Language Models for
  Low-Resource Languages (LoResLM)}, pages 50--61, Rabat, Morocco.

\bibitem[{Hagan(2026)}]{stanfordlegaldesignlab_list}
Margaret Hagan. 2026.
\newblock \href {https://doi.org/10.2139/ssrn.6388659} {You say potato, we say
  legal issue: Developing the legal issues taxonomy (list) for smarter
  issue-spotting and access to justice infrastructure}.
\newblock SSRN.
\newblock Available at SSRN: https://ssrn.com/abstract=6388659.

\bibitem[{Hsia et~al.(2025)Hsia, Shaikh, Wang, and Neubig}]{hsia2024ragged}
Jennifer Hsia, Afreen Shaikh, Zhiruo Wang, and Graham Neubig. 2025.
\newblock \href {https://proceedings.mlr.press/v267/hsia25a.html} {{RAGGED}:
  Towards informed design of scalable and stable {RAG} systems}.
\newblock In \emph{Proceedings of the 42nd International Conference on Machine
  Learning}, volume 267 of \emph{Proceedings of Machine Learning Research},
  pages 24139--24155.

\bibitem[{Hu et~al.(2025)Hu, Liu, Gao, Huang, Han, Feng, Deng, and
  Huang}]{hu2025crosslingual}
Peng Hu, Sizhe Liu, Changjiang Gao, Xin Huang, Xue Han, Junlan Feng, Chao Deng,
  and Shujian Huang. 2025.
\newblock \href {https://doi.org/10.18653/v1/2025.naacl-long.72} {Large
  language models are cross-lingual knowledge-free reasoners}.
\newblock In \emph{Proceedings of the North American Chapter of the Association
  for Computational Linguistics (NAACL)}, pages 1525--1542, Albuquerque, New
  Mexico.

\bibitem[{Kanerva et~al.(2025)Kanerva, Ledins, K{\"a}pyaho, and
  Ginter}]{kanerva2025ocr}
Jenna Kanerva, Cassandra Ledins, Siiri K{\"a}pyaho, and Filip Ginter. 2025.
\newblock \href {https://aclanthology.org/2025.resourceful-1.8/} {{OCR} error
  post-correction with {LLMs} in historical documents: No free lunches}.
\newblock In \emph{Proceedings of the Third Workshop on Resources and
  Representations for Under-Resourced Languages and Domains (RESOURCEFUL)},
  pages 38--47, Tallinn, Estonia. University of Tartu Library.

\bibitem[{Longpre et~al.(2021)Longpre, Perisetla, Chen, Ramesh, DuBois, and
  Singh}]{longpre2021entity}
Shayne Longpre, Kartik Perisetla, Anthony Chen, Nikhil Ramesh, Chris DuBois,
  and Sameer Singh. 2021.
\newblock \href {https://doi.org/10.18653/v1/2021.emnlp-main.565} {Entity-based
  knowledge conflicts in question answering}.
\newblock In \emph{Proceedings of the 2021 Conference on Empirical Methods in
  Natural Language Processing (EMNLP)}, pages 7052--7063.

\bibitem[{Magesh et~al.(2025)Magesh, Surani, Dahl, Suzgun, Manning, and
  Ho}]{magesh2025hallucination}
Varun Magesh, Faiz Surani, Matthew Dahl, Mirac Suzgun, Christopher~D. Manning,
  and Daniel~E. Ho. 2025.
\newblock \href {https://doi.org/10.1111/jels.12413} {Hallucination-free?
  assessing the reliability of leading {AI} legal research tools}.
\newblock \emph{Journal of Empirical Legal Studies}, 22(2):216--242.

\bibitem[{Mohan et~al.(2026)Mohan, Jayakumar, Menon, Kurup, G, and
  Kanjirangat}]{mohan2026retrieval}
Varalekshmy~M Mohan, Swathi Jayakumar, Gadha~Saji Menon, Sachin Kurup, Veena G,
  and Vani Kanjirangat. 2026.
\newblock \href {https://aclanthology.org/2026.mellm-1.9/} {When retrieval
  hurts: Evidence utilization, script fidelity, and knowledge conflicts in
  multilingual {RAG}}.
\newblock In \emph{Proceedings of the 1st Workshop on Multilinguality in the
  Era of Large Language Models (MeLLM)}, pages 92--107, San Diego, United
  States.

\bibitem[{Narayanan and Aepli(2024)}]{madasamy2024tulu}
Manu Narayanan and No{\"e}mi Aepli. 2024.
\newblock \href {https://aclanthology.org/2024.lrec-main.155/} {A tulu resource
  for machine translation}.
\newblock In \emph{Proceedings of the 2024 Joint International Conference on
  Computational Linguistics, Language Resources and Evaluation (LREC-COLING)},
  pages 1756--1767, Torino, Italia. ELRA and ICCL.

\bibitem[{Qi et~al.(2025)Qi, Fern{\'a}ndez, and Bisazza}]{qi2025consistency}
Jirui Qi, Raquel Fern{\'a}ndez, and Arianna Bisazza. 2025.
\newblock \href {https://doi.org/10.18653/v1/2025.mrl-main.15} {On the
  consistency of multilingual context utilization in {RAG}}.
\newblock In \emph{Proceedings of the 5th Workshop on Multilingual
  Representation Learning (MRL)}, pages 199--225, Suzhou, China.
\newblock Best Paper Award.

\bibitem[{Reuter et~al.(2025)Reuter, Lingenberg, Liepi{\c{n}}a, Lagioia, Lippi,
  Sartor, Passerini, and Sayin}]{reuter2025towards}
Markus Reuter, Tobias Lingenberg, R{\=u}ta Liepi{\c{n}}a, Francesca Lagioia,
  Marco Lippi, Giovanni Sartor, Andrea Passerini, and Burcu Sayin. 2025.
\newblock \href {https://doi.org/10.18653/v1/2025.nllp-1.3} {Towards reliable
  retrieval in {RAG} systems for large legal datasets}.
\newblock In \emph{Proceedings of the Natural Legal Language Processing
  Workshop (NLLP)}, pages 17--30, Suzhou, China.

\bibitem[{{Sarvam AI}(2024)}]{sarvamai_models}
{Sarvam AI}. 2024.
\newblock Sarvam-1: A small yet powerful {LLM} for indic languages.
\newblock \url{https://www.sarvam.ai/blogs/sarvam-1}.

\bibitem[{Sun et~al.(2026)Sun, Zhan, Cheang, Wu, Liu, Niu, Ye, Lan, Chao, and
  Wong}]{sun2026exposing}
Yanming Sun, Runzhe Zhan, Chi~Seng Cheang, Han Wu, Xuebo Liu, Yuyao Niu,
  Fengying Ye, Kaixin Lan, Lidia~S. Chao, and Derek~F. Wong. 2026.
\newblock \href {https://doi.org/10.1609/aaai.v40i39.40597} {Exposing the
  cracks: Vulnerabilities of retrieval-augmented {LLM}-based machine
  translation}.
\newblock In \emph{Proceedings of the AAAI Conference on Artificial
  Intelligence}, volume~40, pages 33135--33143.

\bibitem[{Wu et~al.(2024)Wu, Wu, and Zou}]{wu2024clasheval}
Kevin Wu, Eric Wu, and James Zou. 2024.
\newblock \href {https://arxiv.org/abs/2404.10198} {Clasheval: Quantifying the
  tug-of-war between an {LLM's} internal prior and external evidence}.
\newblock In \emph{Advances in Neural Information Processing Systems (NeurIPS),
  Datasets and Benchmarks Track}.

\bibitem[{Xie et~al.(2024)Xie, Zhang, Chen, Lou, and Su}]{xie2024adaptive}
Jian Xie, Kai Zhang, Jiangjie Chen, Renze Lou, and Yu~Su. 2024.
\newblock \href {https://arxiv.org/abs/2305.13300} {Adaptive chameleon or
  stubborn sloth: Revealing the behavior of large language models in knowledge
  conflicts}.
\newblock In \emph{International Conference on Learning Representations
  (ICLR)}.
\newblock Spotlight.

\bibitem[{Xu et~al.(2024)Xu, Qi, Guo, Wang, Wang, Zhang, and
  Xu}]{xu2024knowledge}
Rongwu Xu, Zehan Qi, Zhijiang Guo, Cunxiang Wang, Hongru Wang, Yue Zhang, and
  Wei Xu. 2024.
\newblock \href {https://doi.org/10.18653/v1/2024.emnlp-main.486} {Knowledge
  conflicts for llms: A survey}.
\newblock In \emph{Proceedings of the 2024 Conference on Empirical Methods in
  Natural Language Processing (EMNLP)}, pages 8541--8565, Miami, Florida.

\bibitem[{Zeng et~al.(2025)Zeng, Gupta, Motwani, Yang, and
  Zhang}]{zeng2025ragguard}
Linda Zeng, Rithwik Gupta, Divij Motwani, Diji Yang, and Yi~Zhang. 2025.
\newblock \href {https://arxiv.org/abs/2502.16101} {Worse than zero-shot? a
  fact-checking dataset for evaluating the robustness of {RAG} against
  misleading retrievals}.
\newblock In \emph{Advances in Neural Information Processing Systems
  (NeurIPS)}.

\end{thebibliography}

\clearpage
\onecolumn
\appendix

\section{Full Query Set and Priming Glossary}
\label{app:data}

\subsection{The 60-Sentence Query Set}
\label{app:queries}

Table~\ref{tab:queries-main} lists all 60 Tulu legal-complaint sentences
used in this study, with their human chosen LIST primary and secondary
categories, English gloss, and Romanized-Tulu form. Table~\ref{tab:queries-script}
gives the corresponding Kannada script form.

\setlength{\tabcolsep}{3pt}
\begin{longtable}{p{0.025\textwidth} p{0.115\textwidth} p{0.115\textwidth} p{0.30\textwidth} p{0.30\textwidth}}
\caption{Full 60-sentence query set: human chosen LIST categories, English gloss, and Romanized Tulu.}
\label{tab:queries-main} \\
\toprule
\textbf{ID} & \textbf{Primary} & \textbf{Secondary} & \textbf{English Gloss} & \textbf{Romanized Tulu} \\
\midrule
\endfirsthead
\multicolumn{5}{c}{\tablename\ \thetable{} -- continued} \\
\toprule
\textbf{ID} & \textbf{Primary} & \textbf{Secondary} & \textbf{English Gloss} & \textbf{Romanized Tulu} \\
\midrule
\endhead
\midrule
\multicolumn{5}{r}{\textit{continued on next page}} \\
\endfoot
\bottomrule
\endlastfoot
1 & Housing & Consumer/\allowbreak Money/\allowbreak Debt & My landlord did not give back my money & enna dhani kasn vapas kortijer. \\
2 & Work/\allowbreak Employment & -- & My employer did not pay me this month & enna dhani i tingolda sanbalan kortijer. \\
3 & Torts/\allowbreak Individuals & -- & Someone broke my fence and did not fix it & era enna belin hal maltd nanla sari maltijer. \\
4 & Consumer/\allowbreak Money/\allowbreak Debt & -- & I gave someone money and they disappeared & yan erega kas kori bokka akulu tappayer. \\
5 & Land/\allowbreak Property & -- & My neighbor took my land & enna baritakulu enna jagan detter. \\
6 & Torts/\allowbreak Individuals & -- & After someone hit me with a stick my hand hurts & era enk badutu aki bokka enna kai bene avondundu. \\
7 & Work/\allowbreak Employment & -- & My boss told me to leave without any reason & enna dhani karana korande yan kelasa budodu pand pander. \\
8 & Consumer/\allowbreak Money/\allowbreak Debt & -- & The person I lent money to is not returning it & yan sala kori bokka enk kas vapas korondijjer. \\
9 & Housing & -- & Someone came into my house without my permission & enna oppige ijjande ero enna illda ulayi batter. \\
10 & Consumer/\allowbreak Money/\allowbreak Debt & -- & They want more money than what we decided & akaleg enkulu nirnaya maltina aidd jasti kas kenonduler. \\
11 & Consumer/\allowbreak Money/\allowbreak Debt & Courts/\allowbreak Legal-System & They ended our agreement before the time we decided & akulu enkulu nirnaya maltina samayadda dunbe oppandan mugipayer. \\
12 & Courts/\allowbreak Legal-System & Consumer/\allowbreak Money/\allowbreak Debt & I signed a paper but did not understand what it said & yan daskat padye, matra kakajid dada baretnd pand gottiji. \\
13 & Consumer/\allowbreak Money/\allowbreak Debt & -- & After I signed the paper they did not do what they agreed to & yan kakajid daskat padi bokka enkulu tirpu maltina vishayan akulu malpere kenondijer. \\
14 & Courts/\allowbreak Legal-System & -- & I don't know which court I should go to & enk vovu nyayalayag podu pand gottiji. \\
15 & Courts/\allowbreak Legal-System & Consumer/\allowbreak Money/\allowbreak Debt & The person who owes me money is in a different place & enadda sala detondina jana bete jagad uller. \\
16 & Torts/\allowbreak Individuals & Housing & My neighbor's tree fell on my house and damaged it & enna baritaklena mara enna illda mitt burda enna illan hal maltnd. \\
17 & Land/\allowbreak Property & Housing & Someone is living on my land without asking and won't leave & ero enna oppige ijjande enna jagad kulondu, itte budda popujer. \\
18 & IP/\allowbreak Identity & -- & Someone is selling their things using my name & ero aklena samann enna pudarda maronduler. \\
19 & Work/\allowbreak Employment & -- & My old employer is stopping me from starting my own work & enna paratta dhani enna svanta vyaparan suru malpere budondijer. \\
20 & Estates/\allowbreak Wills & Land/\allowbreak Property & After my father passed away my family took the land that was meant to be given to me & enna popa tir poyi bokka enk barpina jagan enna kutunbadakulu detonyer. \\
21 & Housing & Consumer/\allowbreak Money/\allowbreak Debt & My landlord is trying to raise my rent before our agreement ends & enklena tirpu maltina samayada dunbu enna dhani badige jasti malpere prayatna maltonduller. \\
22 & Housing & Torts/\allowbreak Individuals & The people next door make so much noise I cannot sleep at night & baritakulu et bobbe padonduller panda enk ratred jeppere avondiji. \\
23 & Housing & -- & My landlord will not fix the broken pipe in my kitchen & enna adipilda hal ayina nallin enna dhani sari malpere tayar ijjer. \\
24 & Housing & -- & I was told to leave my rented house with only two days notice & enk badigeda illan budiyere khali radd dinata vayide koryer. \\
25 & Housing & Consumer/\allowbreak Money/\allowbreak Debt & My roommate moved out and left me to pay all the rent & enna sahavasi ill buddu badadigen pura enna mitte padd poyer. \\
26 & Consumer/\allowbreak Money/\allowbreak Debt & -- & The shop owner sold me a broken phone and won't give my money back & enk onji hal ayna phon-n mard daangadida dhani kasn pira korere tayar ijjer. \\
27 & Consumer/\allowbreak Money/\allowbreak Debt & -- & I paid for a service that was never completed & yan pura kas kori bokkala belen purti maltijer. \\
28 & Consumer/\allowbreak Money/\allowbreak Debt & -- & My brother borrowed money from me two years ago and still has not paid & enna megge radd varshada dunbu ennadda sara dettonde, aven nanla pira kortije. \\
29 & Consumer/\allowbreak Money/\allowbreak Debt & -- & The bank is charging me fees I never agreed to & enna oppandina shulkag byank enna mitt kas padondundu. \\
30 & Consumer/\allowbreak Money/\allowbreak Debt & -- & Someone convinced me to invest in something that turned out to be fake & ero enan onji dongi astida mitt kas padere oppayer. \\
31 & Work/\allowbreak Employment & -- & My workplace has not paid me for the extra hours I worked & yan maltina jasti kelsag enk beleda jagadd kas tikkiji. \\
32 & Work/\allowbreak Employment & -- & I was fired the same week I told my boss I was sick & yan dhanita yan hushar ijji pandd pandina varade ar enan beledd detter. \\
33 & Work/\allowbreak Employment & -- & My employer is making me work without the safety equipment I was promised & enna dhani enan suraksha dipina saman korper pand pandi bokka aven korande enada bele malpavonduller. \\
34 & Work/\allowbreak Employment & -- & I was promised a promotion but someone else got it without explanation & enk onji promoshan korper pandd patera koryer, matra dala karana korande bete janak koryer. \\
35 & Work/\allowbreak Employment & Consumer/\allowbreak Money/\allowbreak Debt & My employer refuses to give me my final paycheck after I quit & yan belen budi bokka enna dhani enna akherida sanbalan korondijjer. \\
36 & Land/\allowbreak Property & -- & My neighbor built a wall that crosses onto my land & enna nerekaredakulu aklena belin enna jagada ullayi mutta kattader. \\
37 & Land/\allowbreak Property & -- & Someone is tilling on land that belongs to my family without asking & enda kenande yera enna kutunbada jagad benni dattonduller. \\
38 & Land/\allowbreak Property & -- & The boundary markers on my property were moved without my knowledge & enna jagada belin yera dett budter. \\
39 & Land/\allowbreak Property & Estates/\allowbreak Wills & My brother is claiming a part of our family's land is his alone & enna palaye enna kutunbada jagada onji bhagan ayena matra pandd panondulle. \\
40 & Land/\allowbreak Property & Torts/\allowbreak Individuals & Someone cut down trees on my property without permission & enda kenande yera enna jagada maran kadter. \\
41 & Torts/\allowbreak Individuals & -- & My neighbor's dog bit my child while we were walking & enkulu nadetonduittinapaga enna baritaklena nayi enna balen cuccind. \\
42 & Torts/\allowbreak Individuals & -- & Someone crashed into my parked vehicle and drove away & ero enklena untondu ittina karg akd poyer. \\
43 & Torts/\allowbreak Individuals & -- & A worker dropped something that injured me & onji beledaye enna mitt saman budpad enk tagnd. \\
44 & Torts/\allowbreak Individuals & -- & Someone spread lies about me that damaged my reputation in the village & hallid yera enna bagge sullu pandd enna pudarn hal malter. \\
45 & Torts/\allowbreak Individuals & Land/\allowbreak Property & My neighbor's cattle keep destroying my crops & enna nerekareda aklena pettalu enna pura bulen tindd hal maltnd. \\
46 & Courts/\allowbreak Legal-System & -- & I received a court notice but I don't understand what it is asking me to do & enk nyayalayada onji kakaji baidnd, matra ait dada panonduller pandd gottiji. \\
47 & Courts/\allowbreak Legal-System & -- & I missed a court date because no one explained when I needed to appear & enk nyayalayag popina tarikh tatnd, dayeg panda enk erela epa e podu pandd pandijer. \\
48 & Courts/\allowbreak Legal-System & Work/\allowbreak Employment & I don't know how to file a complaint against my employer & enk enna kelsada dhanina mitt enca duru korodu pandd gottiji. \\
49 & Courts/\allowbreak Legal-System & -- & The police would not register my complaint & polis-da akulu enna durun dettonere tayar ijjer. \\
50 & Courts/\allowbreak Legal-System & -- & I was told I need a lawyer but I don't know how to find one & enk onji vakil bodu pandd pander, matra enk vakiln enca nadodu pandd gottiji. \\
51 & IP/\allowbreak Identity & Consumer/\allowbreak Money/\allowbreak Debt & Someone opened a loan in my name without my knowledge & enk gottijande yera enna pudard sala dettonyer. \\
52 & IP/\allowbreak Identity & -- & My photo is being used in an advertisement without my permission & enada kenande yera enna photon jahiratud upayoga maltonduller. \\
53 & IP/\allowbreak Identity & -- & Someone is pretending to be me online to trick my relatives & yera ennaleka anlain-d natane maltd enna kutunbadaklen dongi maltonduller. \\
54 & IP/\allowbreak Identity & -- & A shop is using my family's business name without asking & vova onji angadi enna kutunbada angadida pudaran upayoga maltondundu \\
55 & IP/\allowbreak Identity & -- & Someone stole the design of the products my family makes and sells & enklena kutunba maltd maruna samanda disain-n yera kandder. \\
56 & Estates/\allowbreak Wills & -- & My father died without a will and my siblings are dividing everything without me & enna popa onji vil budande tir poyi bokka enna megge-palayenakulu enan buddu astin pura pattonduller. \\
57 & Estates/\allowbreak Wills & -- & My mother's jewelry was promised to me but my brother took it after she died & enna ammana bangar enk korper pand pandi bokka amma tir poyi bokka enna palaye aven pura dettonye. \\
58 & Estates/\allowbreak Wills & Land/\allowbreak Property & I found out my parents' house was sold after my father's death without our agreement & enna popa tir poyi bokka enna oppige ijjande enna popa-ammana illan maryer pandd gottand. \\
59 & Estates/\allowbreak Wills & Land/\allowbreak Property & My uncle is refusing to share the property left by my grandfather & enna mamu, ajjer-na jagan pattere kenondijjer. \\
60 & Estates/\allowbreak Wills & Housing & After my husband passed away his family is not letting me stay in our house & enna kandani tir poyi bokka enna kandanina kutunbadakulu enan enklena illad kulyere budondijjer. \\
\end{longtable}

\setlength{\tabcolsep}{6pt}

\renewcommand{\arraystretch}{1.15}
\begin{longtable}{p{0.03\textwidth} >{\kannadafont}p{0.90\textwidth}}
\caption{Full 60-sentence query set in native script (Kannada).}
\label{tab:queries-script} \\
\toprule
\textbf{ID} & \normalfont\textbf{Kannada} \\
\midrule
\endfirsthead
\multicolumn{2}{c}{\tablename\ \thetable{} -- continued} \\
\toprule
\textbf{ID} & \normalfont\textbf{Kannada} \\
\midrule
\endhead
\midrule
\multicolumn{2}{r}{\textit{continued on next page}} \\
\endfoot
\bottomrule
\endlastfoot
1 & ಎನ್ನ ಧನಿ ಕಾಸ್ನ್ ವಾಪಸ್ ಕೊರ್ತಿಜೇರ್. \\
2 & ಎನ್ನ ಧನಿ ಈ ತಿಂಗೋಲ್ದ ಸಂಬಳನ್ ಕೊರ್ತಿಜೇರ್. \\
3 & ಏರಾ ಎನ್ನ ಬೇಲಿನ್ ಹಾಳ್ ಮಲ್ತ್ದ್ ನನ್ಲಾ ಸರಿ ಮಲ್ತಿಜೇರ್. \\
4 & ಯಾನ್ ಏರೇಗಾ ಕಾಸ್ ಕೊರಿ ಬೊಕ್ಕ ಅಕುಲು ತಪ್ಪಾಯೇರ್. \\
5 & ಎನ್ನ ಬರೀತಕುಲು ಎನ್ನ ಜಾಗನ್ ದೆತ್ತೇರ್. \\
6 & ಏರಾ ಎಂಕ್ ಬಡುಟು ಆಕಿ ಬೊಕ್ಕ ಎನ್ನ ಕೈ ಬೇನೆ ಆವೊಂದುಂಡು. \\
7 & ಎನ್ನ ಧನಿ ಕಾರಣ ಕೊರಂದೆ ಯಾನ್ ಕೆಲಸ ಬುಡೊಡು ಪಂದ್ ಪಂಡೆರ್. \\
8 & ಯಾನ್ ಸಾಲ ಕೊರಿ ಬೊಕ್ಕ ಎಂಕ್ ಕಾಸ್ ವಾಪಾಸ್ ಕೊರೋಂದಿಜ್ಜೆರ್. \\
9 & ಎನ್ನ ಓಪ್ಪಿಗೆ ಇಜ್ಜಂದೆ ಏರೋ ಎನ್ನ ಇಲ್ಲ್ದ ಉಲಯಿ ಬತ್ತೇರ್. \\
10 & ಅಕಲೆಗ್ ಎಂಕುಲು ನಿರ್ಣಯ ಮಲ್ತಿನ ಐಡ್ದ್ ಜಾಸ್ತಿ ಕಾಸ್ ಕೇನೋಂದುಲೇರ್. \\
11 & ಅಕುಲು ಎಂಕುಲು ನಿರ್ಣಯ ಮಲ್ತಿನ ಸಮಯಡ್ದ ದುಂಬೇ ಓಪ್ಪಂದನ್ ಮುಗಿಪಾಯೇರ್. \\
12 & ಯಾನ್ ದಸ್ಕತ್ ಪಾಡ್ಯೇ, ಮಾತ್ರ ಕಾಕಜಿಡ್ ದಾದಾ ಬರೆತ್ಂಡ್ ಪಂದ್ ಗೊತ್ತಿಜಿ. \\
13 & ಯಾನ್ ಕಾಕಜಿಡ್ ದಸ್ಕತ್ ಪಾಡಿ ಬೊಕ್ಕ ಎಂಕುಲು ತಿರ್ಪು ಮಲ್ತಿನ ವಿಷಯನ್ ಅಕುಲು ಮಲ್ಪೇರೆ ಕೇನೋಂದಿಜೇರ್. \\
14 & ಎಂಕ್ ವೋವೂ ನ್ಯಾಯಾಲಯಗ್ ಪೋಡು ಪಂದ್ ಗೊತ್ತಿಜಿ. \\
15 & ಎನಡ್ದ ಸಾಲ ದೆತೋಂದಿನ ಜನ ಬೇತೆ ಜಾಗಡ್ ಉಲ್ಲೇರ್. \\
16 & ಎನ್ನ ಬರೀತಕ್ಲೇನ ಮರ ಎನ್ನ ಇಲ್ಲ್ದ ಮಿತ್ತ್ ಬೂರ್ದ ಎನ್ನ ಇಲ್ಲನ್ ಹಾಳ್ ಮಲ್ತ್ಂಡ್. \\
17 & ಏರೋ ಎನ್ನ ಓಪ್ಪಿಗೆ ಇಜ್ಜಂದೆ ಎನ್ನ ಜಾಗಡ್ ಕುಲೋಂದು, ಇತ್ತೇ ಬುಡ್ದ ಪೋಪುಜೇರ್. \\
18 & ಏರೋ ಅಕ್ಲೇನ ಸಾಮಾನ್ನ್ ಎನ್ನ ಪುದರ್ಡ ಮಾರೋಂದುಲೇರ್. \\
19 & ಎನ್ನ ಪರಾತ್ತ ಧನಿ ಎನ್ನ ಸ್ವಂತ ವ್ಯಾಪಾರನ್ ಸುರು ಮಲ್ಪೆರೆ ಬುಡೋಂದಿಜೇರ್. \\
20 & ಎನ್ನ ಪೋಪ ತೀರ್ ಪೋಯಿ ಬೊಕ್ಕ ಎಂಕ್ ಬರ್ಪಿನ ಜಾಗನ್ ಎನ್ನ ಕುಟುಂಬದಕುಲು ದೇತೋಣ್ಯೇರ್. \\
21 & ಎಂಕ್ಲೇನ ತಿರ್ಪು ಮಲ್ತಿನ ಸಮಯದ ದುಂಬು ಎನ್ನ ಧನಿ ಬಾಡಿಗೆ ಜಾಸ್ತಿ ಮಲ್ಪೆರೆ ಪ್ರಯತ್ನ ಮಲ್ತೋಂದುಲ್ಲೇರ್. \\
22 & ಬರೀತಕುಲು ಏತ್ ಬೊಬ್ಬೆ ಪಾಡೋಂದುಲ್ಲೇರ್ ಪನ್ಡ ಎಂಕ್ ರಾತ್ರೆಡ್ ಜೆಪ್ಪೆರೇ ಆವೊಂದಿಜಿ. \\
23 & ಎನ್ನ ಅಡಿಪಿಲ್ದಾ ಹಾಳ್ ಆಯಿನ ನಳ್ಳಿನ್ ಎನ್ನ ಧನಿ ಸರಿ ಮಲ್ಪೆರೆ ತಯಾರ್ ಇಜ್ಜೆರ್. \\
24 & ಎಂಕ್ ಬಾಡಿಗೆದ ಇಲ್ಲನ್ ಬುಡಿಯೆರೆ ಖಾಲಿ ರಡ್ಡ್ ದಿನತ ವಯಿದೇ ಕೊರ್ಯೇರ್. \\
25 & ಎನ್ನ ಸಹವಾಸಿ ಇಲ್ಲ್ ಬುಡ್ದು ಬಾದಡಿಗೆನ್ ಪುರಾ ಎನ್ನ ಮಿತ್ತೆ ಪಾಡ್ದ್ ಪೋಯೇರ್. \\
26 & ಎಂಕ್ ಒಂಜಿ ಹಾಳ್ ಆಯ್ನ ಫೋನ್-ನ್ ಮಾರ್ದ್ ದಅಂಗಡಿದ ಧನಿ ಕಾಸ್ನ್ ಪಿರಾ ಕೊರೆರೆ ತಯಾರ್ ಇಜ್ಜೆರ್. \\
27 & ಯಾನ್ ಪೂರಾ ಕಾಸ್ ಕೊರಿ ಬೊಕ್ಕಲಾ ಬೇಲೆನ್ ಪೂರ್ತಿ ಮಲ್ತಿಜೆರ್. \\
28 & ಎನ್ನ ಮೆಗ್ಗೆ ರಡ್ಡ್ ವರ್ಷದ ದುಂಬು ಎನ್ನಡ್ದ ಸಾರ ದೆತ್ತೊಂಡೆ, ಅವೆನ್ ನನ್ಲಾ ಪಿರಾ ಕೊರ್ತಿಜೆ. \\
29 & ಎನ್ನ ಒಪ್ಪಂದಿನ ಶುಲ್ಕಗ್ ಬ್ಯಾಂಕ್ ಎನ್ನ ಮಿತ್ತ್ ಕಾಸ್ ಪಾಡೊಂದುಂಡು. \\
30 & ಏರೊ ಎನನ್ ಒಂಜಿ ಡೋಂಗಿ ಆಸ್ತಿದ ಮಿತ್ತ್ ಕಾಸ್ ಪಾಡೆರೆ ಒಪ್ಪಾಯೆರ್. \\
31 & ಯಾನ್ ಮಲ್ತಿನ ಜಾಸ್ತಿ ಕೆಲ್ಸಗ್ ಎಂಕ್ ಬೇಲೆದ ಜಾಗಡ್ಡ್ ಕಾಸ್ ತಿಕ್ಕಿಜಿ. \\
32 & ಯಾನ್ ಧನಿಟ ಯಾನ್ ಹುಷಾರ್ ಇಜ್ಜಿ ಪಂಡ್ಡ್ ಪಂಡಿನ ವಾರಡೆ ಆರ್ ಎನನ್ ಬೇಲೆಡ್ಡ್ ದೆತ್ತೆರ್. \\
33 & ಎನ್ನ ಧನಿ ಎನನ್ ಸುರಕ್ಷ ದೀಪಿನ ಸಾಮಾನ್ ಕೊರ್ಪೆರ್ ಪಂದ್ ಪಂಡಿ ಬೊಕ್ಕ ಅವೆನ್ ಕೊರಂದೇ ಎನಡ ಬೇಲೆ ಮಲ್ಪಾವೊಂದುಲ್ಲೆರ್. \\
34 & ಎಂಕ್ ಒಂಜಿ ಪ್ರೊಮೋಷನ್ ಕೊರ್ಪೆರ್ ಪಂಡ್ಡ್ ಪಾತೇರ ಕೊರ್ಯೆರ್, ಮಾತ್ರ ದಾಲ ಕಾರಣ ಕೊರಂದೇ ಬೇತೆ ಜನಕ್ ಕೊರ್ಯೆರ್. \\
35 & ಯಾನ್ ಬೇಲೆನ್ ಬುಡಿ ಬೊಕ್ಕ ಎನ್ನ ಧನಿ ಎನ್ನ ಅಖೇರಿದ ಸಂಬಳನ್ ಕೊರೋಂದಿಜ್ಜೆರ್. \\
36 & ಎನ್ನ ನೆರೆಕರೆದಕುಲು ಅಕ್ಲೆನ ಬೇಲಿನ್ ಎನ್ನ ಜಾಗದ ಉಲ್ಲಯಿ ಮುಟ್ಟ ಕಟ್ಟಾದೆರ್. \\
37 & ಎನ್ಡ ಕೇಣಂದೇ ಯೇರಾ ಎನ್ನ ಕುಟುಂಬದ ಜಾಗಡ್ ಬೆನ್ನಿ ದಟ್ಟೊಂದುಲ್ಲೆರ್. \\
38 & ಎನ್ನ ಜಾಗದ ಬೇಲಿನ್ ಯೇರಾ ದೆತ್ತ್ ಬುಡ್ತೆರ್. \\
39 & ಎನ್ನ ಪಲಯೆ ಎನ್ನ ಕುಟುಂಬದ ಜಾಗದ ಒಂಜಿ ಭಾಗನ್ ಆಯೇನ ಮಾತ್ರ ಪಂಡ್ಡ್ ಪಣೊಂದುಲ್ಲೆ. \\
40 & ಎನ್ಡ ಕೇಣಂದೇ ಯೇರಾ ಎನ್ನ ಜಾಗದ ಮರನ್ ಕಡ್ತೆರ್. \\
41 & ಎಂಕುಲು ನಡೆತೊಂದುಿತ್ತಿನಪಗ ಎನ್ನ ಬರಿತ್ಅಕ್ಲೆನ ನಾಯಿ ಎನ್ನ ಬಾಲೆನ್ ಚೂಚ್ಚಿಂಡ್. \\
42 & ಏರೊ ಎಂಕ್ಲೆನ ಉಂತೋಂದು ಇತ್ತಿನ ಕಾರ್ಗ್ ಆಕ್ದ್ ಪೋಯೆರ್. \\
43 & ಒಂಜಿ ಬೇಲೆದಾಯೆ ಎನ್ನ ಮಿತ್ತ್ ಸಾಮಾನ್ ಬೂಡ್ಪಾದ್ ಎಂಕ್ ತಾಗ್ಂಡ್. \\
44 & ಹಳ್ಳಿಡ್ ಯೇರಾ ಎನ್ನ ಬಾಗ್ಗೆ ಸುಳ್ಳು ಪಂಡ್ಡ್ ಎನ್ನ ಪುದರ್ನ್ ಹಾಳ್ ಮಲ್ತೆರ್. \\
45 & ಎನ್ನ ನೆರೆಕರೆದ ಅಕ್ಲೆನ ಪೆತ್ತಲು ಎನ್ನ ಪೂರಾ ಬುಳೆನ್ ತಿಂದ್ಡ್ ಹಾಳ್ ಮಲ್ತ್ಂಡ್. \\
46 & ಎಂಕ್ ನ್ಯಾಯಾಲಯದ ಒಂಜಿ ಕಾಕಜಿ ಬೈದ್ಂಡ್, ಮಾತ್ರ ಐಟ್ ದಾದ ಪಣೊಂದುಲ್ಲೆರ್ ಪಂಡ್ಡ್ ಗೊತ್ತಿಜಿ. \\
47 & ಎಂಕ್ ನ್ಯಾಯಾಲಯಗ್ ಪೋಪಿನ ತಾರೀಖ್ ತತ್ಂಡ್, ದಾಯೆಗ್ ಪಂಡ ಎಂಕ್ ಏರೇಲಾ ಏಪ ಏ ಪೋಡು ಪಂಡ್ಡ್ ಪಂಡಿಜೆರ್. \\
48 & ಎಂಕ್ ಎನ್ನ ಕೆಲ್ಸದ ಧನಿನ ಮಿತ್ತ್ ಎಂಚ ದೂರು ಕೊರೋಡು ಪಂಡ್ಡ್ ಗೊತ್ತಿಜಿ. \\
49 & ಪೊಲೀಸ್-ದ ಅಕುಲು ಎನ್ನ ದೂರುನ್ ದೆತ್ತೊನೆರೆ ತಯಾರ್ ಇಜ್ಜೆರ್. \\
50 & ಎಂಕ್ ಒಂಜಿ ವಕೀಲ್ ಬೋಡು ಪಂಡ್ಡ್ ಪಂಡೆರ್, ಮಾತ್ರ ಎಂಕ್ ವಕೀಲ್ನ್ ಎಂಚ ನಾಡೋಡು ಪಂಡ್ಡ್ ಗೊತ್ತಿಜಿ. \\
51 & ಎಂಕ್ ಗೊತ್ತಿಜಂದೇ ಯೇರಾ ಎನ್ನ ಪುದರ್ಡ್ ಸಾಲ ದೆತ್ತೊನ್ಯೆರ್. \\
52 & ಎನಡ ಕೇನಂದೇ ಯೇರಾ ಎನ್ನ ಫೋಟೋನ್ ಜಾಹೀರಾತುಡ್ ಉಪಯೋಗ ಮಲ್ತೊಂದುಲ್ಲೆರ್. \\
53 & ಯೇರಾ ಎನ್ನಲೆಕ ಆನ್‌ಲೈನ್-ಡ್ ನಟನೆ ಮಲ್ತ್ಡ್ ಎನ್ನ ಕುಟುಂಬದಕ್ಲೆನ್ ಡೋಂಗಿ ಮಲ್ತೊಂದುಲ್ಲೆರ್. \\
54 & ವೊವಾ ಒಂಜಿ ಅಂಗಡಿ ಎನ್ನ ಕುಟುಂಬದ ಅಂಗಡಿದ ಪುದರನ್ ಉಪಯೋಗ ಮಲ್ತೊಂದುಂಡು \\
55 & ಎಂಕ್ಲೆನ ಕುಟುಂಬ ಮಲ್ತ್ಡ್ ಮಾರುನ ಸಾಮಾಂದ ಡಿಸೈನ್-ನ್ ಯೇರಾ ಕಂಡ್ದೆರ್. \\
56 & ಎನ್ನ ಪೋಪ ಒಂಜಿ ವಿಲ್ ಬುಡಂದೇ ತೀರ್ ಪೋಯಿ ಬೊಕ್ಕ ಎನ್ನ ಮೆಗ್ಗೆ-ಪಲಯೆನಕುಲು ಎನನ್ ಬುಡ್ಡು ಆಸ್ತಿನ್ ಪೂರಾ ಪಟ್ಟೋಂದುಲ್ಲೆರ್. \\
57 & ಎನ್ನ ಅಮ್ಮನ ಬಂಗಾರ್ ಎಂಕ್ ಕೊರ್ಪೆರ್ ಪಂಡ್ ಪಂಡಿ ಬೊಕ್ಕ ಅಮ್ಮ ತೀರ್ ಪೋಯಿ ಬೊಕ್ಕ ಎನ್ನ ಪಲಯೆ ಅವೆನ್ ಪೂರಾ ದೆತ್ತೊನ್ಯೆ. \\
58 & ಎನ್ನ ಪೋಪ ತೀರ್ ಪೋಯಿ ಬೊಕ್ಕ ಎನ್ನ ಒಪ್ಪಿಗೆ ಇಜ್ಜಂದೆ ಎನ್ನ ಪೋಪ-ಅಮ್ಮನ ಇಲ್ಲನ್ ಮಾರ್ಯೆರ್ ಪಂಡ್ಡ್ ಗೊತ್ತಾಂಡ್. \\
59 & ಎನ್ನ ಮಾಮು, ಅಜ್ಜೆರ್-ನ ಜಾಗನ್ ಪಟ್ಟೆರೆ ಕೇನೋಂದಿಜ್ಜೆರ್. \\
60 & ಎನ್ನ ಕಂಡನಿ ತೀರ್ ಪೋಯಿ ಬೊಕ್ಕ ಎನ್ನ ಕಂಡನಿನ ಕುಟುಂಬದಕುಲು ಎನನ್ ಎಂಕ್ಲೆನ ಇಲ್ಲಡ್ ಕುಲ್ಯೆರೆ ಬುಡೋಂದಿಜ್ಜೆರ್. \\
\end{longtable}

\renewcommand{\arraystretch}{1}

\subsection{Priming Glossary}
\label{app:glossary}

The priming glossary (Table~\ref{tab:gloss-safe} and Table~\ref{tab:gloss-leaky})
lists the vocabulary items used to construct the sparse and extensive
priming blocks (\S\ref{subsec:experimental-conditions}). \textbf{Safe-tier} items are grammatical/functional
vocabulary with no category-revealing content; \textbf{leaky-tier} items
are content words (e.g.\ ``landlord,'' ``money,'' ``land'') that overlap
with terms likely to appear in specific LIST categories and are the
vocabulary most responsible for the jump in accuracy under extensive
priming (\S\ref{subsec:comprehension_transfer}). 

\begin{longtable}{>{\kannadafont}p{0.20\textwidth} p{0.20\textwidth} p{0.45\textwidth}}
\caption{Safe-tier glossary items (grammatical/function vocabulary, no category-revealing content).}
\label{tab:gloss-safe} \\
\toprule
\normalfont\textbf{Kannada} & \textbf{Romanized} & \textbf{Meaning} \\
\midrule
\endfirsthead
\multicolumn{3}{c}{\tablename\ \thetable{} -- continued} \\
\toprule
\normalfont\textbf{Kannada} & \textbf{Romanized} & \textbf{Meaning} \\
\midrule
\endhead
\bottomrule
\endlastfoot
ಆಂಜೊ & aanjo & man \\
ಆಲ್ & al & she \\
ಬಾಲೆ & baale & child \\
ಮೀನ್ & meen & fish \\
ಪೋಪಿನಿ & popini & to go \\
ಮಲ್ಪುನಿ & malpuni & to do/make \\
ಕೊರ್ಪುನಿ & korpuni & to give \\
ಗೊತ್ತು & gothu & to know \\
ಬೋಡು & bodu & to want \\
ವಾಪಸ್ & wapas & return/back \\
ಮುಗ್ಪಾವುನಿ & mugpavuni & to end/finish \\
ಇಜ್ಜಿ & ijji & negation(not) \\
ಅತ್ತ್ & att & negative \\
ಬೊಕ್ಕ & boca & after \\
ದುಂಬು & dumbu & before \\
ಮಾತ್ರ & matra & but/contrast \\
ಪಂಡ & panda & quotative(THAT) \\
ಉಂಡು & undu & exists/is \\
ಅಮ್ಮ & amma & mother \\
\end{longtable}

\begin{longtable}{>{\kannadafont}p{0.20\textwidth} p{0.20\textwidth} p{0.45\textwidth}}
\caption{Leaky-tier glossary items (content vocabulary overlapping with LIST category terms).}
\label{tab:gloss-leaky} \\
\toprule
\normalfont\textbf{Kannada} & \textbf{Romanized} & \textbf{Meaning} \\
\midrule
\endfirsthead
\multicolumn{3}{c}{\tablename\ \thetable{} -- continued} \\
\toprule
\normalfont\textbf{Kannada} & \textbf{Romanized} & \textbf{Meaning} \\
\midrule
\endhead
\bottomrule
\endlastfoot
ಯಜಮಾನ & yejamaana & landlord/owner \\
ಧನಿ & dhani & landlord/employer (authority figure over you) \\
ಕಾಸು & kas & money \\
ಜಾಗ & jagga & land \\
ಇಲ್ಲ್ & ille & house \\
ಬೇಲಿ & belee & fence \\
ನೆರೆಕರೆ & nerakare & neighbor \\
ಪುದರ್ & pudar & name \\
ಕಾಕಜಿ & kakaji & document \\
ಕುಟುಂಬ & kotumba & family \\
ಸಾಲ & sala & loan/debt \\
ಓಪ್ಪಿಗೆ & oppigay & permission \\
\end{longtable}

\twocolumn

\section{Prompt Template}
\label{app:prompt}

Every model call in this study, across all 19 conditions and all three
models, was assembled by a single unified prompt-builder function shared
across the whole pipeline (\texttt{build\_unified\_prompt}), so that script,
priming tier, and RAG inclusion could vary independently without a separate,
drifting prompt implementation for each condition. Each call is fully
stateless: no conversation history or KV-cache continuity carries over
between queries (Section~\ref{sec:mitigation}). The full prompt sent to the
model is assembled from up to five parts, always in this order:

\begin{enumerate}
\item A \textbf{static prefix}, identical across all 19 conditions: system
      role, LIST category definitions, an anti-collapse instruction, and two
      few-shot calibration examples.
\item An optional \textbf{priming block} (sparse- or full-tier conditions
      only), giving a script-specific Tulu vocabulary guide.
\item The \textbf{situation block}: the query itself, with a script-specific
      framing sentence.
\item For RAG conditions only, up to three \textbf{retrieved passages} from
      the Kannada court corpus.
\item \textbf{Output instructions}: a three-step reasoning scaffold ending in
      a top-3 JSON prediction.
\end{enumerate}

\subsection{Static Prefix (all 19 conditions)}
\label{app:prompt-prefix}

\begin{quote}
\ttfamily\small
You are a legal AI assistant helping classify everyday legal situations described by lay speakers into legal issue categories.

The legal categories are:

- Housing: landlord-tenant disputes, security deposits not returned, trespass into a home or dwelling, damage to rented property, eviction, someone entering your house without permission. Examples: `My landlord did not give back my deposit', `Someone came into my house without asking'.

- Consumer/Money/Debt: money owed but not paid, loans not repaid, overcharging beyond what was agreed, contract breaches involving payment, fraud where money was taken and the person disappeared. Examples: `I lent money and they won't return it', `They charged more than we agreed', `I gave someone money and they disappeared', `They ended our agreement early'.

- Work/Employment: wages not paid, wrongful dismissal without reason, employer stopping you from starting your own business, non-compete situations. Examples: `My employer did not pay me this month', `My boss told me to leave without any reason', `My old employer is stopping me from starting my own work'.

- Land/Property: neighbor taking your land, someone occupying your land without permission and refusing to leave, boundary disputes, land stolen after inheritance. Examples: `My neighbor took my land', `Someone is living on my land without asking and won't leave'.

- Torts/Individuals: personal injury caused by another person, property damage caused by another person's negligence, someone breaking your property and not fixing it, a neighbor's tree falling on your house. Examples: `Someone hit me with a stick and my hand hurts', `Someone broke my fence and did not fix it', `My neighbor's tree fell on my house'.

- Courts/Legal-System: confusion about which court to go to, not knowing where to file a complaint, the other party being in a different location or jurisdiction, signing a document without understanding what it said. Examples: `I don't know which court I should go to', `The person who owes me money is in a different place', `I signed a paper but did not understand what it said'.

- IP/Identity: someone using your name without permission, someone selling things under your name or brand, misuse of personal or business identity. Examples: `Someone is selling their things using my name'.

- Estates/Wills: inheritance disputes after a family member's death, family members taking property that was meant to be given to you, land or assets disputed after someone passes away. Examples: `After my father passed away my family took the land that was meant for me'.

IMPORTANT: These categories describe EVERYDAY PROBLEMS, not commercial contract clauses. Classify based on what the PERSON'S SITUATION is about, NOT based on legal terminology in retrieved passages.

Do not default to the same category repeatedly. Weigh each situation against ALL categories before deciding.

Here are two example classifications to calibrate your reasoning:

Example 1:
Situation: "My landlord kept my deposit money after I moved out."
Classification reasoning: The person had a rental relationship (Housing) and the landlord failed to return money (Consumer/Money/Debt). The primary issue is the landlord-tenant relationship and deposit -- Housing.
\{"category": "Housing", "confidence": 0.85, "reasoning": "Landlord-tenant deposit dispute"\}

Example 2:
Situation: "Someone broke my fence and refused to fix it."
Classification reasoning: This is damage caused by another individual to my property. It is not a housing/landlord issue (no tenancy). It is not a debt (no money owed from an agreement). It is physical harm/damage caused by another person -- Torts/Individuals.
\{"category": "Torts/Individuals", "confidence": 0.80, "reasoning": "Property damage caused by another person's action"\}
\end{quote}

\subsection{Priming Block (sparse/full-tier conditions only)}
\label{app:prompt-priming}

For conditions with a priming tier, this block is appended immediately after
the static prefix. Sparse tier draws only from the ``safe'' (grammatical
function vocabulary) items in the glossary (Appendix~\ref{app:glossary});
full tier draws from the entire glossary, safe and leaky tiers combined. The
template (script-agnostic: \texttt{\{SCRIPT\}} and each
\texttt{word=meaning} pair change per script and tier):

\begin{quote}
\ttfamily\small
Vocabulary guide for Tulu in \{SCRIPT\} script:\\
Nouns: \{word=meaning, word=meaning, ...\}\\
Legal nouns: \{word=meaning, ...\} \quad\textit{(full tier only)}\\
Verbs: \{word=meaning, ...\}\\
Grammar: \{word=meaning, ...\}\\
SOV word order -- verb always at end. Suffix -g/-k = `to/towards'. Suffix -n/-d = `of/belonging to'. Quotative marks embedded speech.
\end{quote}

\subsection{Situation Block}
\label{app:prompt-situation}

The framing sentence changes by script; the closing instruction changes by
whether a priming block was included.

\begin{quote}
\ttfamily\small
\textbf{English:} SITUATION (in English): "\{display\_text\}"\\[4pt]
\textbf{Romanized:} SITUATION (Tulu, romanized): "\{display\_text\}"\\[4pt]
\textbf{Kannada:} SITUATION (Tulu written in Kannada script): "\{display\_text\}"\\[8pt]
\textit{-- closing instruction, with priming:}\\
No translation available. Use the vocabulary guide above and any words you recognize in the query to understand the situation.\\[4pt]
\textit{-- closing instruction, without priming:}\\
No translation available. Use any patterns you can detect in the text.
\end{quote}

\subsection{Retrieved-Passage Block (RAG conditions only)}
\label{app:prompt-rag}

\begin{quote}
\ttfamily\small
The passages below are BACKGROUND LAW ONLY. Do NOT classify what the passage is about -- classify what the PERSON'S SITUATION is about.\\
Retrieved passages from the Kannada court corpus:\\[4pt]
Passage 1 (similarity: \{score\}, source: \{case\_name\}, LIST categories for this case: \{labels\}):\\
\{passage text\}\\[4pt]
Passage 2 ...\\
Passage 3 ...
\end{quote}

\noindent Passage text and scores shown here are template placeholders, not
a fabricated example; see the methodological note in
Appendix~\ref{app:rag-examples} regarding retrieved-passage preservation for
this batch.

\subsection{Output Instructions (all conditions)}
\label{app:prompt-output}

\begin{quote}
\ttfamily\small
Step 1: Re-read the SITUATION above. In one sentence, describe in plain English what problem this person is facing -- ignore the passages for now.\\
Step 2: Based on YOUR description of the situation (not the passages), identify the best-fitting LIST category and explain why. Rule out at least two other categories explicitly.\\
Step 3: Provide your TOP 3 category predictions in order, as JSON:
{\footnotesize
\begin{verbatim}
```json
{
  "top1": {
    "category": "...",
    "confidence": 0.0-1.0,
    "reasoning": "..."
  },
  "top2": {
    "category": "...",
    "confidence": 0.0-1.0,
    "reasoning": "..."
  },
  "top3": {
    "category": "...",
    "confidence": 0.0-1.0,
    "reasoning": "..."
  }
}
```
\end{verbatim}
}
CRITICAL: Every category value MUST be exactly one of:\\
\ \ - Housing\\
\ \ - Consumer/Money/Debt\\
\ \ - Work/Employment\\
\ \ - Land/Property\\
\ \ - Torts/Individuals\\
\ \ - Courts/Legal-System\\
\ \ - IP/Identity\\
\ \ - Estates/Wills\\
Do NOT invent new category names or use terms from the passages.
\end{quote}

\subsection{Complete Worked Example}
\label{app:prompt-worked}

To ground the template above in an actual assembled call rather than
placeholders, this is the complete, real prompt for Condition E\_full
(Kannada script, full priming, no RAG) on query~1 (human chosen label: Housing) --
static prefix (Appendix~\ref{app:prompt-prefix}) followed by the full-tier
Kannada priming block (built from the real glossary in
Appendix~\ref{app:glossary}) and situation block, exactly as sent to the
model:

\begin{quote}
\ttfamily\small
[\,...static prefix as in Appendix~\ref{app:prompt-prefix}...\,]

Vocabulary guide for Tulu in Kannada script:\\
Nouns: {\kannadafont ಆಂಜೊ}=man, {\kannadafont ಆಲ್}=she, {\kannadafont ಬಾಲೆ}=child, {\kannadafont ಮೀನ್}=fish, {\kannadafont ಅಮ್ಮ}=mother\\
Legal nouns: {\kannadafont ಯಜಮಾನ}=landlord/owner, {\kannadafont ಧನಿ}=landlord/employer (authority figure over you), {\kannadafont ಕಾಸು}=money, {\kannadafont ಜಾಗ}=land, {\kannadafont ಇಲ್ಲ್}=house, {\kannadafont ಬೇಲಿ}=fence, {\kannadafont ನೆರೆಕರೆ}=neighbor, {\kannadafont ಪುದರ್}=name, {\kannadafont ಕಾಕಜಿ}=document, {\kannadafont ಕುಟುಂಬ}=family, {\kannadafont ಸಾಲ}=loan/debt, {\kannadafont ಓಪ್ಪಿಗೆ}=permission, {\kannadafont ದಸ್ಕತ್}=signature\\
Verbs: {\kannadafont ಪೋಪಿನಿ}=to go, {\kannadafont ಮಲ್ಪುನಿ}=to do/make, {\kannadafont ಕೊರ್ಪುನಿ}=to give, {\kannadafont ಗೊತ್ತು}=to know, {\kannadafont ಬೋಡು}=to want, {\kannadafont ವಾಪಸ್}=return/back, {\kannadafont ಮುಗ್ಪಾವುನಿ}=to end/finish\\
Grammar: {\kannadafont ಇಜ್ಜಿ}=negation(not), {\kannadafont ಅತ್ತ್}=negative, {\kannadafont ಬೊಕ್ಕ}=after, {\kannadafont ದುಂಬು}=before, {\kannadafont ಮಾತ್ರ}=but/contrast, {\kannadafont ಪಂಡ}=quotative(THAT), {\kannadafont ಉಂಡು}=exists/is\\
SOV word order -- verb always at end. Suffix -g/-k = `to/towards'. Suffix -n/-d = `of/belonging to'. Quotative marks embedded speech.

\medskip
\textbf{--- CLASSIFY THIS SITUATION ---}

SITUATION (Tulu written in Kannada script): "{\kannadafont ಎನ್ನ ಧನಿ ಕಾಸ್ನ್ ವಾಪಸ್ ಕೊರ್ತಿಜೇರ್.}"\\
No translation available. Use the vocabulary guide above and any words you recognize in the query to understand the situation.

No retrieved passages are provided for this classification.\\
Classify based solely on the situation description and the category definitions above.

REMINDER -- classify this situation:\\
SITUATION (Tulu written in Kannada script): "{\kannadafont ಎನ್ನ ಧನಿ ಕಾಸ್ನ್ ವಾಪಸ್ ಕೊರ್ತಿಜೇರ್.}"\\
No translation available. Use the vocabulary guide above and any words you recognize in the query to understand the situation.

[\,...output instructions as in Appendix~\ref{app:prompt-output}...\,]
\end{quote}

\section{Representative RAG Reasoning-Trace Examples}
\label{app:rag-examples}


\subsection{Fact Substitution}
\label{app:factsub}

\begin{quote}
\textbf{Model:} Llama3 \quad \textbf{Condition:} B (Romanized Tulu, no priming, RAG) \\
\textbf{Query (gloss):} ``Someone cut down trees on my property without permission'' \\
\textbf{Human chosen label:} Land/Property \\
\quad \textbf{Predicted:} Consumer/Money/Debt \\
\textbf{Reasoning trace:} ``The situation appears to be related to a
financial issue, specifically a debt or payment problem, given the
mention of insufficient funds and a check being returned.'' \\
\textbf{Verification:} The third retrieved passage was drawn from
\textit{Sridhar P.\ v.\ Ramesh}, a cheque-bounce dispute containing
Section 138 Negotiable Instruments Act language and explicit references
to a returned cheque ({\kannadafont ಚೆಕ್}). Neither cheques nor money are
mentioned anywhere in the query, which concerns unauthorized tree-felling
on the speaker's land; the retrieved passage's content was substituted
directly into the reasoning in place of the query's actual content.
\end{quote}

\begin{quote}
\textbf{Model:} Sarvam \quad \textbf{Condition:} C-sparse (Romanized Tulu, sparse priming, RAG) \\
\textbf{Query (gloss):} ``My roommate moved out and left me to pay all the rent'' \\
\textbf{Human chosen label:} Housing \\
\quad \textbf{Predicted:} Courts/Legal-System \\
\textbf{Reasoning trace:} ``The person is actively involved in a legal
process, mentioning filing a document, producing evidence (WhatsApp
screenshots), and undergoing a court procedure (fingerprinting).'' \\
\textbf{Verification:} The first retrieved passage
(\textit{Amith Ramachandra v.\ Shantha}) contains both details verbatim:
{\kannadafont ವಾಟ್ಸಾಪ್ ಸಂದೇಶಗಳ ಸ್ಕ್ರೀನ್ ಶಾಟ್} (WhatsApp message
screenshots), and ಬೆರಳಚ್ಚು (typing, which the model misunderstood as fingerprinting).  Nothing resembling either detail appears in the
query (a roommate/rent dispute).
\end{quote}

\subsection{Confabulation}
\label{app:confab}

\begin{quote}
\textbf{Model:} Llama3 \quad \textbf{Condition:} D-full (Kannada script, full priming, RAG) \\
\textbf{Query (gloss):} ``The people next door make so much noise I cannot sleep at night'' \\
\textbf{Human chosen label:} Housing \\
\quad \textbf{Predicted:} Courts/Legal-System \\
\textbf{Reasoning trace:} ``The situation involves threats and demands,
which suggests that legal action may be involved or necessary to resolve
the issue.'' \\
\textbf{Verification:} Neither ``threats'' nor ``demands'' (nor Kannada
equivalents {\kannadafont ಬೆದರಿಕೆ} / {\kannadafont ಬೇಡಿಕೆ}) appear in the
query or in any of the three retrieved passages (\textit{Amith
Ramachandra v.\ Shantha}; \textit{Sadashiva Shetty v.\ Vijaya Bank};
\textit{State v.\ Sudhakar Pai}) (though threats appear in the \textit{Amith
Ramachandra v.\ Shantha} text in a seperate chunk) . The reasoning invents a threat scenario
with no basis anywhere in the available context.
\end{quote}

\begin{quote}
\textbf{Model:} Llama3 \quad \textbf{Condition:} F (Kannada script, no priming, RAG) \\
\textbf{Query (gloss):} ``My neighbor's dog bit my child while we were walking'' \\
\textbf{Human chosen label:} Torts/Individuals \quad \textbf{Predicted:} Estates/Wills \\
\textbf{Reasoning trace:} ``The situation involves eviction and family \\
members, which suggests an issue related to property or inheritance.'' \\
\textbf{Verification:} No eviction-related term
({\kannadafont ಖಾಲಿ}, {\kannadafont ಉಚ್ಚಾಟನೆ}) appears in the query or in
any of the three retrieved passages (\textit{Nikitha A.\ v.\ Santhosh
Ghatke}; \textit{Akkamma v.\ Ramesh Patil}; \textit{Dr.\ Alfred Aranha
v.\ Sahel Peter Aranha}). ``Eviction'' is introduced with no basis in
either source, though family members are brought up in both \textit{Nikitha A.\ v.\ Santhosh Ghatke} and \textit{Akkamma v.\ Ramesh Patil}.
\end{quote}

\subsection{RAG Content and Correct Classification Aligning}
\label{app:aligning}

\begin{quote}
\textbf{Model:} Sarvam \quad \textbf{Condition:} F (Kannada script, no priming, RAG) \\
\textbf{Query (gloss):} ``My neighbor's dog bit my child while we were walking'' \\
\textbf{Human chosen label:} Torts/Individuals \\
\quad \textbf{Predicted:} Torts/Individuals \\
\textbf{Reasoning trace:} ``This is a personal injury case caused by another person's (or their animal's) negligence or action, which falls under the Torts/Individuals category.'' \\
\textbf{Verification:} The retrieved passage (\textit{Nikitha A.\ v.\ Santhosh Ghatke}, similarity 0.916) is actually on track with Torts/Individuals content. Llama3 confabulates on this same query under the same condition (see above), predicting Estates/Wills with reasoning citing ``eviction and family members''. The relevance of a retrieved passage does not ensure accurate classification.
\end{quote}

\begin{quote}
\textbf{Model:} Hex-1 \quad \textbf{Condition:} A (English gloss, RAG) \\
\textbf{Query (gloss):} ``After my father passed away my family took the land that was meant to be given to me'' \\
\textbf{Human chosen label:} Estates/Wills\\
\quad \textbf{Predicted:} Estates/Wills \\
\textbf{Reasoning trace:} ``The situation involves inheritance disputes
after a family member's death.'' \\
\textbf{Verification:} The retrieved passage (\textit{Gagan B.S.\ v.\
Shivamurthy T.}, similarity 0.824) is itself an ancestral-property
inheritance-share dispute ({\kannadafont ಪಿತ್ರಾಜರ್ಿತ
ಆಸ್ತಿ}, ``ancestral property''). Retrieval and classification align
correctly here as well.
\end{quote}

\subsection{Relevant Retrieval, Persistent Failure}
\label{app:relevantfail}

\begin{quote}
\textbf{Model:} Sarvam \quad \textbf{Condition:} D-full (Kannada script, full priming, RAG) \\
\textbf{Query (gloss):} ``They ended our agreement before the time we decided'' \\
\textbf{Human chosen label:} Consumer/Money/Debt\\ 
\quad \textbf{Predicted:} Land/Property \\
\textbf{Reasoning trace (translated):} ``This situation involves a
contract dispute related to property rights or control. This is more of
a property and contract dispute than a landlord-tenant relationship.'' \\
\textbf{Verification:} The retrieved passage (\textit{Sridhar P.\ v.\
Ramesh}, similarity 0.911) is an on-topic financial/debt
dispute, though financial matters are discussed briefly. Despite having retrieved directly relevant material, the model still produced an unrelated category.
\end{quote}

\section{Full Condition-Level Results}
\label{app:full-findings}

\begin{figure*}[t]
\centering
\includegraphics[width=\linewidth]{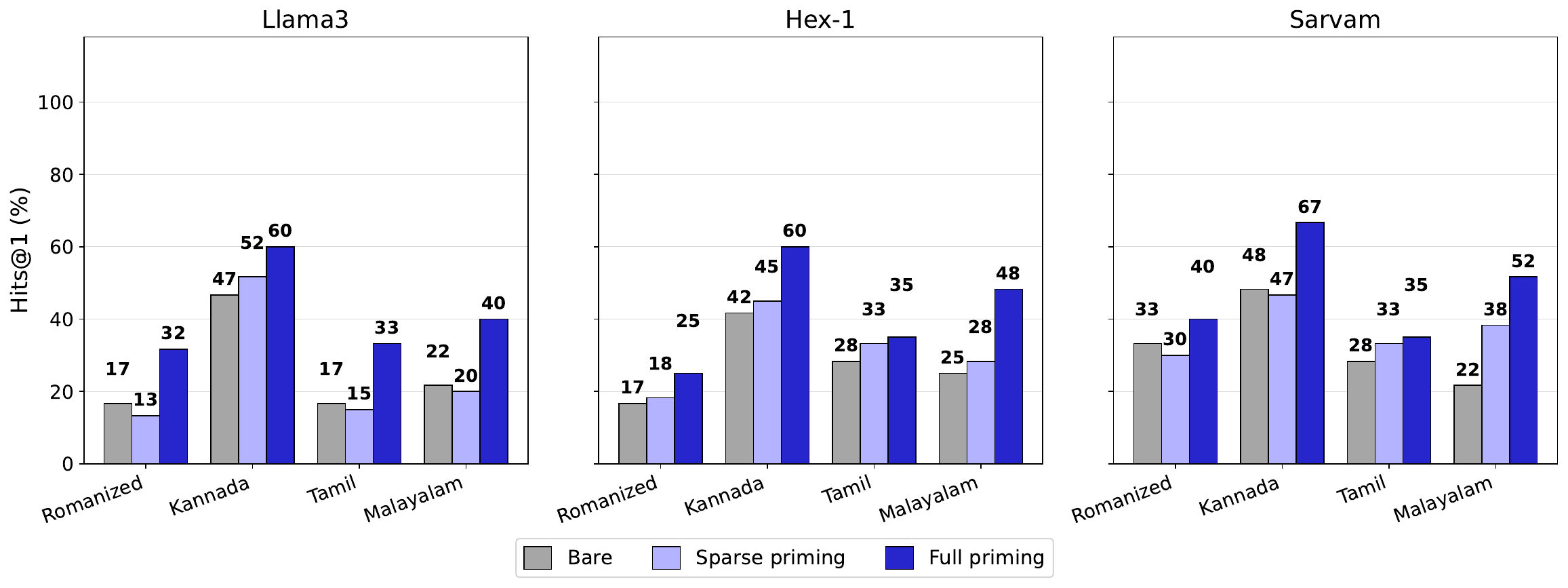}
\caption{Hits@1 by script and priming tier, per model ($n=60$/bar).}
\label{fig:script-priming}
\end{figure*}

\begin{figure*}[t]
\centering
\includegraphics[width=1\linewidth]{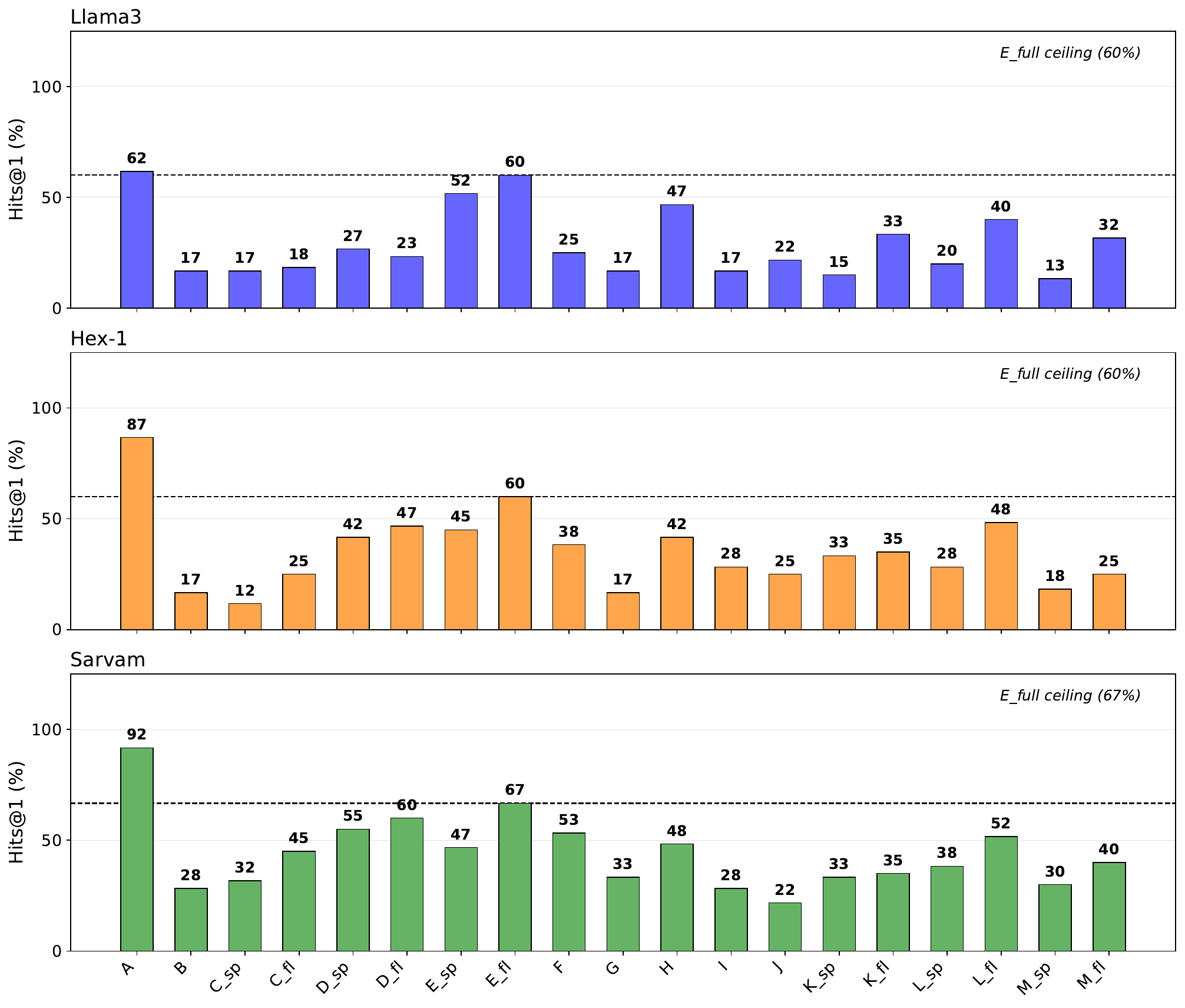}
\caption{Hits@1 across all 19 conditions, per model ($n=60$/bar). Each panel's dashed line marks that model's own no-RAG, full-priming ceiling (E\_full).}
\label{fig:a-h-comparison}
\end{figure*}

This appendix gives the complete comprehension and RAG results underlying the condensed Findings in \S\ref{sec:findings}: the full script $\times$ priming-tier breakdown (Table~\ref{tab:comprehension}, Figure~\ref{fig:script-priming}), the full 19-condition breakdown (Table~\ref{tab:rag-results}, Figure~\ref{fig:a-h-comparison}), and the complete significance test battery (Table~\ref{tab:significance-summary}).

\begin{table*}[t]
\centering
\small
\begin{tabular}{@{}llcccccc@{}}
\toprule
& & \multicolumn{2}{c}{\textbf{Llama3}} & \multicolumn{2}{c}{\textbf{Hex-1}} & \multicolumn{2}{c}{\textbf{Sarvam}} \\
\cmidrule(lr){3-4}\cmidrule(lr){5-6}\cmidrule(lr){7-8}
\textbf{Script} & \textbf{Tier} & \textbf{Hits@1} & \textbf{Hits@3} & \textbf{Hits@1} & \textbf{Hits@3} & \textbf{Hits@1} & \textbf{Hits@3} \\
\midrule
\multirow{3}{*}{Romanized} & bare   & 16.7\% & 46.7\% & 16.7\% & 65.0\% & 33.3\% & 61.7\% \\
                            & sparse & 13.3\% & 43.3\% & 18.3\% & 55.0\% & 30.0\% & 60.0\% \\
                            & full   & 31.7\% & 56.7\% & 25.0\% & 61.7\% & 40.0\% & 71.7\% \\
\midrule
\multirow{3}{*}{Kannada} & bare   & 46.7\% & 76.7\% & 41.7\% & 68.3\% & 48.3\% & 76.7\% \\
                          & sparse & 51.7\% & 75.0\% & 45.0\% & 76.7\% & 46.7\% & 78.3\% \\
                          & full   & 60.0\% & 83.3\% & 60.0\% & 80.0\% & 66.7\% & 88.3\% \\
\midrule
\multirow{3}{*}{Tamil} & bare   & 16.7\% & 48.3\% & 28.3\% & 66.7\% & 28.3\% & 55.0\% \\
                        & sparse & 15.0\% & 51.7\% & 33.3\% & 61.7\% & 33.3\% & 56.7\% \\
                        & full   & 33.3\% & 73.3\% & 35.0\% & 65.0\% & 35.0\% & 68.3\% \\
\midrule
\multirow{3}{*}{Malayalam} & bare   & 21.7\% & 53.3\% & 25.0\% & 60.0\% & 21.7\% & 51.7\% \\
                            & sparse & 20.0\% & 55.0\% & 28.3\% & 68.3\% & 38.3\% & 68.3\% \\
                            & full   & 40.0\% & 76.7\% & 48.3\% & 73.3\% & 51.7\% & 76.7\% \\
\bottomrule
\end{tabular}
\caption{Comprehension accuracy by script and priming tier, no RAG, all three models ($n=60$/cell).}
\label{tab:comprehension}
\end{table*}

\begin{table*}[t]
\centering
\small
\begin{tabular}{@{}lccccccc@{}}
\toprule
\textbf{Condition} & \textbf{Llama3 H1} & \textbf{Llama3 H3} & \textbf{Hex-1 H1} & \textbf{Hex-1 H3} & \textbf{Sarvam H1} & \textbf{Sarvam H3} & \textbf{n} \\
\midrule
A         & 61.7\% & 95.0\% & 86.7\% & 98.3\% & 91.7\% & 100.0\% & 60 \\
B         & 16.7\% & 56.7\% & 16.7\% & 50.0\% & 28.3\% & 63.3\%  & 60 \\
C\_sparse & 16.7\% & 46.7\% & 11.7\% & 46.7\% & 31.7\% & 61.7\%  & 60 \\
C\_full   & 18.3\% & 53.3\% & 25.0\% & 63.3\% & 45.0\% & 73.3\%  & 60 \\
D\_sparse & 26.7\% & 66.7\% & 41.7\% & 68.3\% & 55.0\% & 75.0\%  & 60 \\
D\_full   & 23.3\% & 68.3\% & 46.7\% & 80.0\% & 60.0\% & 80.0\%  & 60 \\
E\_sparse & 51.7\% & 75.0\% & 45.0\% & 76.7\% & 46.7\% & 78.3\%  & 60 \\
E\_full   & 60.0\% & 83.3\% & 60.0\% & 80.0\% & 66.7\% & 88.3\%  & 60 \\
F         & 25.0\% & 60.0\% & 38.3\% & 71.7\% & 53.3\% & 71.7\%  & 60 \\
G         & 16.7\% & 46.7\% & 16.7\% & 65.0\% & 33.3\% & 61.7\%  & 60 \\
H         & 46.7\% & 76.7\% & 41.7\% & 68.3\% & 48.3\% & 76.7\%  & 60 \\
I         & 16.7\% & 48.3\% & 28.3\% & 66.7\% & 28.3\% & 55.0\%  & 60 \\
J         & 21.7\% & 53.3\% & 25.0\% & 60.0\% & 21.7\% & 51.7\%  & 60 \\
K\_sparse & 15.0\% & 51.7\% & 33.3\% & 61.7\% & 33.3\% & 56.7\%  & 60 \\
K\_full   & 33.3\% & 73.3\% & 35.0\% & 65.0\% & 35.0\% & 68.3\%  & 60 \\
L\_sparse & 20.0\% & 55.0\% & 28.3\% & 68.3\% & 38.3\% & 68.3\%  & 60 \\
L\_full   & 40.0\% & 76.7\% & 48.3\% & 73.3\% & 51.7\% & 76.7\%  & 60 \\
M\_sparse & 13.3\% & 43.3\% & 18.3\% & 55.0\% & 30.0\% & 60.0\%  & 60 \\
M\_full   & 31.7\% & 56.7\% & 25.0\% & 61.7\% & 40.0\% & 71.7\%  & 60 \\
\bottomrule
\end{tabular}
\caption{Hits@1 / Hits@3 across all 19 conditions, per model ($n=60$ each).}
\label{tab:rag-results}
\end{table*}

\begin{table*}[t]
\centering
\small
\begin{tabular}{@{}p{6.4cm}ccc@{}}
\toprule
\textbf{Comparison (Hits@1)} & \textbf{Llama3} & \textbf{Hex-1} & \textbf{Sarvam} \\
\midrule
\multicolumn{4}{@{}l}{\textit{Baseline comprehension (binomial test vs.\ chance/majority)}} \\
H (Kannada bare) vs.\ chance (12.5\%) / majority (25\%) & $p<.001$ / $p<.001$ & $p<.001$ / $p<.001$ & $p<.001$ / $p<.001$ \\
D\_full vs.\ majority-class (25\%)                & $p=.665$ (ns)       & $p<.001$***       & $p<.001$*** \\
\midrule
\multicolumn{4}{@{}l}{\textit{Priming effects (Fisher's exact)}} \\
E\_full vs.\ H (Kannada full-priming effect)      & $p=.200$ (ns)       & $p=.067$ (ns)       & $p=.064$ (ns) \\
L\_full vs.\ J (Malayalam full-priming effect)    & $p=.047$*           & $p=.013$*           & $p=.001$** \\
\midrule
\multicolumn{4}{@{}l}{\textit{RAG effects (Fisher's exact)}} \\
H vs.\ F (RAG collapse, Kannada bare)             & $p=.022$*           & $p=.852$ (ns)       & $p=.715$ (ns) \\
E\_sparse vs.\ D\_sparse (RAG collapse, Kannada sparse) & $p=.009$**    & $p=.854$ (ns)       & $p=.465$ (ns) \\
E\_full vs.\ D\_full (RAG collapse, Kannada full) & $p<.001$***         & $p=.200$ (ns)       & $p=.570$ (ns) \\
A vs.\ D\_full (query language under RAG)         & $p<.001$***         & $p<.001$***         & $p<.001$*** \\
\bottomrule
\end{tabular}
\caption{Fisher's exact and binomial tests for pairwise condition comparisons at Hits@1, $n=60$/condition. Grouped by comparison type: baseline comprehension checks against chance/majority guessing, priming-effect checks (no RAG), and RAG-effect checks (Kannada bare, sparse, and full priming, plus query-language effect under RAG). $^{*}p<.05$, $^{**}p<.01$, $^{***}p<.001$.}
\label{tab:significance-summary}
\end{table*}

\begin{table*}[t]
\centering
\small
\begin{tabular}{lcccccc}
\toprule
\textbf{Model} & \textbf{Bare} & \textbf{Sparse prime} & \textbf{Full prime} & \textbf{RAG only} & \textbf{RAG+sparse} & \textbf{RAG+full} \\
\midrule
Llama3 & 46.7>16.7*** & 51.7>13.3*** & 60.0>31.7** & 25.0>16.7 (ns) & 26.7>16.7 (ns) & 23.3>18.3 (ns) \\
Hex-1  & 41.7>16.7**  & 45.0>18.3**  & 60.0>25.0*** & 38.3>16.7*     & 41.7>11.7***   & 46.7>25.0* \\
Sarvam & 48.3>33.3 (ns) & 46.7>30.0 (ns) & 66.7>40.0** & 53.3>28.3**  & 55.0>31.7*     & 60.0>45.0 (ns) \\
\bottomrule
\end{tabular}
\caption{Kannada-script vs. Romanized-script comprehension (Hits@1, \%) across all six priming/RAG tiers, per model ($n=60$/condition). Values shown as Kannada\% $>$ Romanized\%. Fisher's exact test, same methodology as Table~\ref{tab:significance-summary}. Computed from condition-level results in Table~\ref{tab:rag-results}; no new model calls were made. $^*p<.05$, $^{**}p<.01$, $^{***}p<.001$.}
\label{tab:kannada-romanized}
\end{table*}

\clearpage
\onecolumn
\begin{table}[tbp]
\centering
\small
\begin{tabular}{llccc}
\toprule
\textbf{Model} & \textbf{Tier} & \textbf{E-right/D-wrong} & \textbf{D-right/E-wrong} & \textbf{Ratio} \\
\midrule
Llama3 & full   & 26 & 4 & 6.5:1 \\
Llama3 & sparse & 20 & 5 & 4.0:1 \\
Hex-1  & full   & 10 & 2 & 5.0:1 \\
Hex-1  & sparse & 8  & 6 & 1.3:1 \\
Sarvam & full   & 7  & 3 & 2.3:1 \\
Sarvam & sparse & 4  & 9 & 0.4:1 \\
\bottomrule
\end{tabular}
\caption{Disagreement-level analysis: among cases where full/sparse-priming no-RAG (E) and RAG (D) predictions diverge, how often each condition was uniquely correct, per model and priming tier ($n=60$/condition).}
\label{tab:disagreement-ratio}
\end{table}

\begin{table}[t]
\centering
\small
\begin{tabular}{lcccc}
\hline
\textbf{Judge language} & \textbf{Hits@1} & \textbf{Hits@3} & \textbf{Passages retained} & \textbf{No-RAG fallback} \\
\hline
Kannada script   & 25.0\% (5/20)  & 65.0\% (13/20) & 2.85 / 3 & 0/20 \\
English gloss    & 60.0\% (12/20) & 80.0\% (16/20) & 1.65 / 3 & 3/20 \\
\hline
\end{tabular}
\caption{CRAG relevance-verification results, Llama3, Kannada-script D-condition queries ($n=20$). Fisher's exact test on Hits@1 (English-judge vs.\ Kannada-judge): $p = .054$. Reference points from the full 60-query set (Table~\ref{tab:rag-results}): D\_full (RAG, full priming) = 23.3\%; E\_full (no-RAG, full priming) = 60.0\%.}
\label{tab:crag}
\end{table}

\section{Court Cases Cited}
\label{app:cases}

Table~\ref{tab:cases-cited} lists full identifiers for the eight court
cases named in Appendix~\ref{app:rag-examples}, drawn from the
18-document Kannada legal corpus (\S\ref{subsec:data-models}). All documents are public
lower-court filings retrieved from 
ecourtsindia.com; CNR is the eCourts Case Number Record, the
portal's unique case identifier.

\begin{table}[H]
\centering
\small
\begin{tabular}{@{}p{3.7cm}p{2.6cm}p{5.3cm}p{2.4cm}@{}}
\toprule
\textbf{Case} & \textbf{Case No.} & \textbf{Court} & \textbf{CNR / Date} \\
\midrule
Sridhar P.\ v.\ Ramesh & C.C.\ No.\ 5343/2024 & 4th Addl.\ Small Causes Court \& ACJM, Bengaluru Urban & KABC020187742024, 3 Jun 2025 \\
Amith Ramachandra v.\ Shantha & P.C.R.\ No.\ 3981/2025 & 41st Addl.\ Chief Judicial Magistrate, Bengaluru Urban & KABC030166202025, 6 May 2025 \\
Sadashiva Shetty v.\ Vijaya Bank & R.A.\ No.\ 43-2001 & Principal Civil Judge (Sr.\ Div.) \& CJM, Udupi & KAUP020002942001, 24 Jan 2009 \\
State v.\ Sudhakar Pai & C.C.\ No.\ 1357/2016 & JMFC 2nd Court, Mangaluru & KADK070014602016, 18 Oct 2016 \\
Nikitha A.\ v.\ Santhosh Ghatke & Crl.Misc.\ No.\ 127/2023 & JMFC Traffic Court VI, Bengaluru Urban & KABC080042862023, 4 Jul 2025 \\
Akkamma v.\ Ramesh Patil & Cri.Misc.\ No.\ 17/2018 & Principal Civil Judge \& JMFC, Byadgi, Haveri & KAHV210000762018, 1 Aug 2018 \\
Dr.\ Alfred Aranha v.\ Sahel Peter Aranha & O.S.\ No.\ 419/2013 & 4th Addl.\ Senior Civil Judge \& 1st Class JMFC, Mangaluru & KADK030000092013, 2 Feb 2019 \\
Gagan B.S.\ v.\ Shivamurthy T. & Property partition suit & Senior Civil Judge \& JMFC, Shikaripura, Shivamogga & KASM510014042020, 16 Aug 2025 \\
\bottomrule
\end{tabular}
\caption{Court cases named in Appendix~\ref{app:rag-examples}, with case number, deciding court, eCourts CNR identifier, and order date. All are public lower-court filings from ecourtsindia.com .}
\label{tab:cases-cited}
\end{table}

\end{document}